%% file: main.tex
\documentclass{article}

     \PassOptionsToPackage{numbers, compress}{natbib}

\usepackage[preprint]{neurips_2021}

\input{math_commands.tex}

\usepackage[utf8]{inputenc} %
\usepackage[T1]{fontenc}    %
\usepackage[colorlinks=true,urlcolor=red]{hyperref}       %
\usepackage{url}            %
\usepackage{booktabs}       %
\usepackage{amsfonts}       %
\usepackage{nicefrac}       %
\usepackage{microtype}      %
\usepackage{xcolor}         %
\usepackage{subfigure}

\usepackage{amsmath}
\usepackage{amssymb} 
\usepackage{algpseudocode}
\usepackage{algorithm}
\usepackage{amsthm}
\usepackage{bm}
\usepackage{multirow}
\usepackage{boldline}
\usepackage{comment}
\usepackage{graphicx}
\usepackage{array}
\usepackage{makecell}
\usepackage{verbatim}
\usepackage{bbm}
\usepackage{verbatim}
\usepackage{caption} 
\usepackage{blindtext}
\usepackage{svg}
\usepackage{enumitem}

\usepackage[title]{appendix}

\usepackage{wrapfig}

\definecolor{mydarkblue}{rgb}{0,0.08,0.45}

\hypersetup{
     colorlinks=true,
     linkcolor=red,
     filecolor=red,
     citecolor = mydarkblue,      
     urlcolor=red,
     }

\makeatletter
\DeclareRobustCommand\onedot{\futurelet\@let@token\@onedot}
\def\@onedot{\ifx\@let@token.\else.\null\fi\xspace}

\def\eg{\emph{e.g}\onedot} 
\def\ie{\emph{i.e}\onedot} 
 
 \def\vs{\emph{vs}\onedot}

\makeatother
\newcommand{\dinfty}[1]{\boldsymbol{\delta}^{(\infty)}}
\newcommand{\dtwo}[1]{\boldsymbol{\delta}^{(2)}}
\newcommand{\done}[1]{\boldsymbol{\delta}^{(1)}}

\title{Robustifying $\ell_\infty$  Adversarial Training to the Union of Perturbation Models}

\author{%
  Ameya D.~Patil, Michael Tuttle, Alexander G. Schwing, and Naresh R. Shanbhag\\
  University of Illinois at Urbana-Champaign\\
  Urbana, IL 61801 \\
  \texttt{\{adpatil2,mtuttle3,aschwing,shanbhag\}@illinois.edu} \\
}

\begin{document}

\maketitle

\begin{abstract}
Classical adversarial training (AT) frameworks are designed to achieve high adversarial accuracy against a single attack type, typically $\ell_\infty$ norm-bounded perturbations. 
Recent extensions in AT have focused on defending against the union of multiple perturbations but this benefit is obtained at the expense of a significant (up to $10\times$) increase in training complexity over single-attack $\ell_\infty$ AT.
In this work, we expand the capabilities of widely popular single-attack $\ell_\infty$ AT frameworks to provide robustness to the union of ($\ell_\infty, \ell_2, \ell_1$) perturbations while preserving their training efficiency.
Our technique, referred to as \textbf{S}haped \textbf{N}oise \textbf{A}ugmented \textbf{P}rocessing (\textbf{SNAP}),
exploits a well-established byproduct of single-attack AT frameworks -- the reduction in the curvature of the decision boundary of networks. SNAP 
prepends a given deep net with a shaped noise augmentation layer whose distribution is learned along with network parameters using any standard single-attack AT. As a result, SNAP enhances adversarial accuracy of ResNet-18 on CIFAR-10 against the union of ($\ell_\infty, \ell_2, \ell_1$) perturbations by $14\%$-to-$20\%$ for four state-of-the-art (SOTA) single-attack $\ell_\infty$ AT frameworks, and, for the first time, establishes a benchmark for ResNet-50 and ResNet-101 on ImageNet.
\end{abstract}

\input{intro}

\input{relatedworks}

\input{insight}

\input{ellinoise}

\input{results}

\input{discussion}

\begin{ack}
This work was supported by the Semiconductor Research Corporation (SRC) and DARPA sponsored Center for Brain-inspired Computing (C-BRIC) and SRC's AIHW program.
\end{ack}

\input{Supplementary}

\end{document}

%% file: math_commands.tex
\usepackage{amsmath,amsfonts,bm}

\def\eqref#1{equation~\ref{#1}}

\def\1{\bm{1}}

\def\rvn{{\mathbf{n}}}

\def\va{{\bm{a}}}

\def\vp{{\bm{p}}}

\def\vs{{\bm{s}}}

\def\vu{{\bm{u}}}
\def\vv{{\bm{v}}}

\def\vx{{\bm{x}}}
\def\vy{{\bm{y}}}

\DeclareMathAlphabet{\mathsfit}{\encodingdefault}{\sfdefault}{m}{sl}
\SetMathAlphabet{\mathsfit}{bold}{\encodingdefault}{\sfdefault}{bx}{n}

%% file: intro.tex
\section{Introduction}
\label{sec:intro}

Today \emph{adversarial training} (AT) provides state-of-the-art (SOTA) empirical defense against adversarial perturbations. %
For this, 
adversarial perturbations are used during training to optimize a \emph{robust} loss function~\citep{madry2018towards,zhang2019theoretically,shafahi2019adversarial,wong2020fast}. %
Early AT frameworks~\cite{madry2018towards,zhang2019theoretically} were $7\times$-to-$10\times$ more computationally demanding than vanilla training. More recent works~\citep{shafahi2019adversarial,wong2020fast,zhang2019you} have significantly reduced the  computational demands of AT via %
\emph{single-step attacks} and \emph{superconvergence}. %

\begin{wrapfigure}[]{r}{0.45\linewidth}
    \vspace{-0.4cm}
	\begin{center}
		\includegraphics[width=\linewidth]{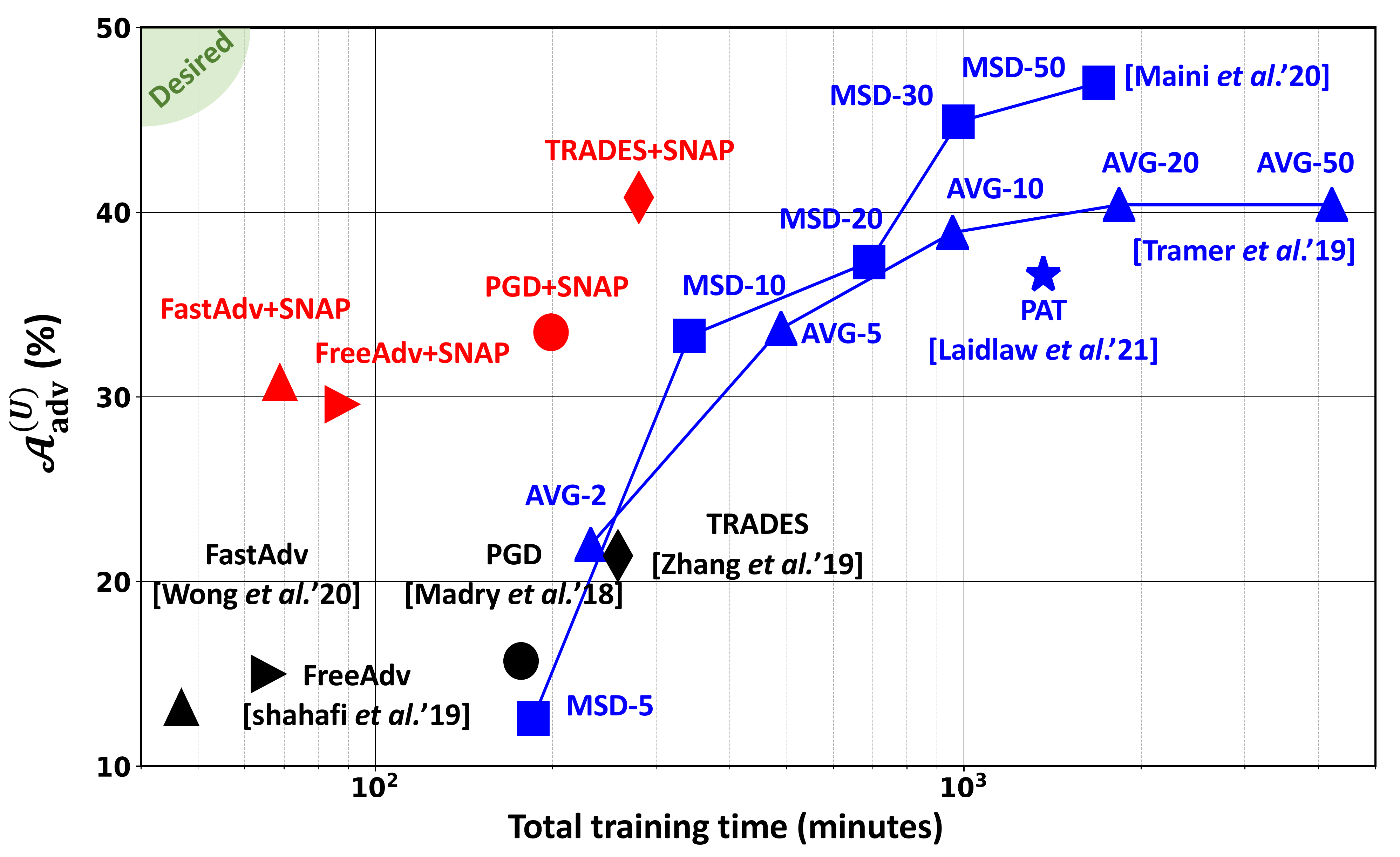}
	\end{center}
	\vspace{-0.4cm}
	\caption{Adversarial accuracy ($\mathcal{A}^{(U)}_\text{adv}$) against union of ($\ell_\infty$, $\ell_2$, $\ell_1$) \vs measured wall-clock total training time on CIFAR-10  with different AT frameworks on single NVIDIA TESLA P100 GPU. $\epsilon=(0.031, 0.5, 12)$ for ($\ell_\infty$, $\ell_2$, $\ell_1$) perturbations, respectively. SNAP enhances robustness with a small increase in training 
	time. All frameworks except PAT employ ResNet-18.}
	\label{fig::IntroFig} 
	\vspace{-0.7cm}
\end{wrapfigure}

However, today's AT frameworks  predominantly focus on a \emph{single-attack}, \ie, they seek robustness to a single  perturbation, typically $\ell_\infty$-bounded~\cite{shafahi2019adversarial,wong2020fast,xie2020intriguing,zhang2019theoretically,zheng2020efficient,zhang2019you,yang2020closer,rebuffi2021fixing,gowal2020uncovering,vivek2020single,zhang2020attacks,gui2019model,guo2020meets,hu2020triple}. This results in low performance against other perturbations such as $\ell_2$, $\ell_1$, or the union of ($\ell_\infty, \ell_2, \ell_1$). Indeed, as shown in Fig.~\ref{fig::IntroFig}, four  state-of-the-art (SOTA) single-attack AT frameworks (\emph{black markers}) employing only $\ell_\infty$-bounded perturbations achieve low adversarial accuracy $\mathcal{A}^{(U)}_\text{adv}$ of $\approx 15\%$-to-$20\%$ against the union of ($\ell_\infty, \ell_2, \ell_1$) perturbations. Recent extensions in AT~\citep{maini2019adversarial,tramer2019adversarial,laidlaw2021perceptual} do seek higher $\mathcal{A}^{(U)}_\text{adv}$ but only at the expense of a  $6\times$-to-$10\times$ increase in the total training time (\emph{blue markers in Fig.~\ref{fig::IntroFig}}). The large training time of these AT frameworks has inhibited their application to large-scale datasets such as ImageNet, \eg, \citet{maini2019adversarial,tramer2019adversarial} show results for MNIST and CIFAR-10 only, while \citet{laidlaw2021perceptual} only additionally show $64\times64$ ImageNet-100 results.

The high training time for AT frameworks arises from two sources: (i) the need to employ larger networks, \eg, MSD~\cite{maini2019adversarial} with ResNet-18 achieves higher $\mathcal{A}^{(U)}_\text{adv}$ than  PAT~\citep{laidlaw2021perceptual} with ResNet-50 (see Fig.~\ref{fig::IntroFig}); and (ii) the need to incorporate multiple perturbations during each attack step and a higher overall number of attack steps, \eg, 50 in MSD~\citep{maini2019adversarial}, 20 in AVG~\citep{tramer2019adversarial}. Obviously one can always reduce the number of attack steps in MSD/AVG to proportionally reduce training time. Doing so results in  training time and $\mathcal{A}^{(U)}_\text{adv}$ to rapidly approach the training complexity and $\mathcal{A}^{(U)}_\text{adv}$ of standard AT frameworks, \eg, a 5-step MSD and 2-step AVG is equivalent in training time and accuracy to PGD and TRADES, respectively. Notwithstanding the expensive nature of 50-step multi-attack training, today MSD~\citep{maini2019adversarial} achieves a SOTA $\mathcal{A}^{(U)}_\text{adv}$ of 47\% with ResNet-18 on CIFAR-10.

This poses a question: can we
approach the high robustness of multiple-attack AT such as 50-step MSD against the union of ($\ell_\infty, \ell_2, \ell_1$) perturbations while maintaining the low training time of fast single-attack AT frameworks such as FreeAdv~\citep{shafahi2019adversarial} and FastAdv~\citep{wong2020fast}? %

In our quest to answer this question we find 
that %
noise augmentation using adequately shaped noise within standard single-attack AT frameworks employing $\ell_\infty$-bounded perturbations %
significantly improves robustness against the union of ($\ell_\infty, \ell_2, \ell_1$) perturbations. The improvement appears to be a  consequence of a well-established byproduct of AT frameworks -- the reduction in the curvature of the decision boundary of networks trained using single-attack AT~\citep{dezfooli2018robustness,moosavi2019robustness}. We confirm this connection by quantifying the impact of single-attack AT on the geometric orientations of different perturbations. 

Based on this insight, we propose  \textbf{S}haped \textbf{N}oise \textbf{A}ugmented \textbf{P}rocessing (\textbf{SNAP}) -- \emph{a method to enhance robustness against the union of perturbation types by augmenting single-attack AT frameworks}. SNAP prepends a deep net with a shaped noise (SN) augmentation layer (see Fig.~\ref{fig::ElliNoiseConcept}) whose distribution parameter $\Sigma$ is learned with that of the network ($\theta$) within any standard single-attack AT framework. SNAP improves the robustness of four SOTA $\ell_\infty$-AT frameworks against the union of ($\ell_\infty, \ell_2, \ell_1$) perturbations by 15\%-to-20\% on CIFAR-10 (\emph{red markers in Fig.~\ref{fig::IntroFig}}) with only a modest ($\sim 10\%$) increase in training time. %
This expands the capabilities of widely popular single-attack $\ell_\infty$ AT frameworks to providing robustness to the union of ($\ell_\infty, \ell_2, \ell_1$) perturbations  without sacrificing  training efficiency. We validate SNAP's benefits via thorough comparisons with \emph{nine SOTA adversarial training and randomized smoothing frameworks} across different operating regimes on both CIFAR-10 and ImageNet. %

One tangible outcome of our work -- we demonstrate \emph{for the first time} ResNet-50 (ResNet-101) networks on ImageNet that achieve $\mathcal{A}^{(U)}_\text{adv}=32\%$ (35\%) against the union of ($\ell_\infty (\epsilon=2/255), \ell_2 (\epsilon = 2.0), \ell_1 (\epsilon = 72.0)$) perturbations. %
Our code is available at \href{https://github.com/adpatil2/SNAP}{https://github.com/adpatil2/SNAP}.

%% file: relatedworks.tex
\vspace{-0.3cm}
\section{Related Work}
\label{sec::RelatedWorks}
\vspace{-0.2cm}
We categorize works on adversarial vulnerability of DNNs as follows:

\noindent{\textbf{Low-complexity adversarial training}:} The high computational needs of AT frameworks has spurred significant efforts in reducing their complexity~\cite{zhang2019you,shafahi2019adversarial,wong2020fast,zheng2020efficient}. FreeAdv \cite{shafahi2019adversarial} updates weights while accumulating multiple attack iterations. FastAdv \cite{wong2020fast} employs \emph{appropriate} use of single-step attacks, while \citet{zheng2020efficient} leverage inter-epoch similarity between adversarial perturbations. However, these fast AT methods seek  robustness against a single perturbation type, \eg,~$\ell_\infty$ norm-bounded perturbations. In contrast, SNAP expands the capabilities of these AT frameworks by enhancing robustness to the union of three perturbation types ($\ell_\infty, \ell_2, \ell_1$), while preserving their efficiency.

\noindent{\textbf{Robustness against union of perturbation models}:} The focus on the robustness against the union of multiple perturbation types is relatively new. \citet{kang2019transfer} studied transferability between different perturbation types, while \citet{jordan2019quantifying} considered combination attacks with low perceptual distortion. \citet{stutz2020confidence} proposed a modification in AT to \emph{detect} images with different models of perturbations via confidence thresholding, but they don't attempt to \emph{classify} perturbed images correctly. For accurate classification in the presence of different perturbation models, \citet{tramer2019adversarial} studied empirical and theoretical trade-offs involved in including multiple perturbation types simultaneously during training. \citet{maini2019adversarial} further built upon this work to propose the multi steepest descent (MSD) AT framework which chooses one among the three perturbation models ($\ell_\infty, \ell_2, \ell_1$) in each attack iteration during training, achieving SOTA adversarial accuracy on CIFAR-10 against the union of the ($\ell_\infty, \ell_2, \ell_1$) perturbation models, albeit at a high ($10\times$) training time. In contrast, SNAP provides high robustness against the union of ($\ell_\infty, \ell_2, \ell_1$) perturbation models using established single-attack $\ell_\infty$ AT frameworks. This enables  to showcase the benefits of our approach on large-scale datasets such as ImageNet.

Recently, \citet{laidlaw2021perceptual} developed a novel AT framework (PAT) with low perceptual distortion attacks to demonstrate impressive generalization to unseen attacks. In contrast, we focus on extending the capabilities of widely popular $\ell_\infty$-AT frameworks to providing robustness against the union of ($\ell_\infty, \ell_2, \ell_1$) perturbations, while preserving their training efficiency. 

\noindent{\textbf{Noise augmentation}:} 
Multiple recent works have investigated the role of randomization in enhancing adversarial robustness \citep{he2019parametric,pinot2019theoretical,gilmer2019adversarial,pinot2020randomization} with theoretical guarantees. %
Another prominent line of work in this category is randomized smoothing \citep{cohen2019certified,salman2019provably,li2019certified,yang2020randomized}, where random noise is used as a tool to compute certification bounds. \citet{rusak2020simple} also explored the role of noise augmentation for improving the robustness against  common-corruptions~\cite{hendrycks2018benchmarking}.  In contrast, in SNAP, noise augmentation is used as a means to enable widely popular $\ell_\infty$-AT frameworks to efficiently achieve high robustness against the union of multiple norm-bounded perturbations. As is the characteristic of AT works, our results are primarily empirical in nature. Hence, we follow recent guidelines \citep{tramer2020adaptive,maini2019adversarial} to evaluate the accuracy against the strongest possible adversaries. We do explicitly compare $\ell_\infty$-AT+SNAP with randomized smoothing approaches in the Appendix~\ref{app::RandSmooth}.

%% file: insight.tex
\vspace{-0.4cm}
\section{Subspace Analysis of Adversarial Perturbations}
\label{sec::SubspaceAnalysis}

\begin{wrapfigure}[]{r}{0.405\linewidth}
    \vspace{-0.6cm}
	\begin{center}
		\includegraphics[width=\linewidth]{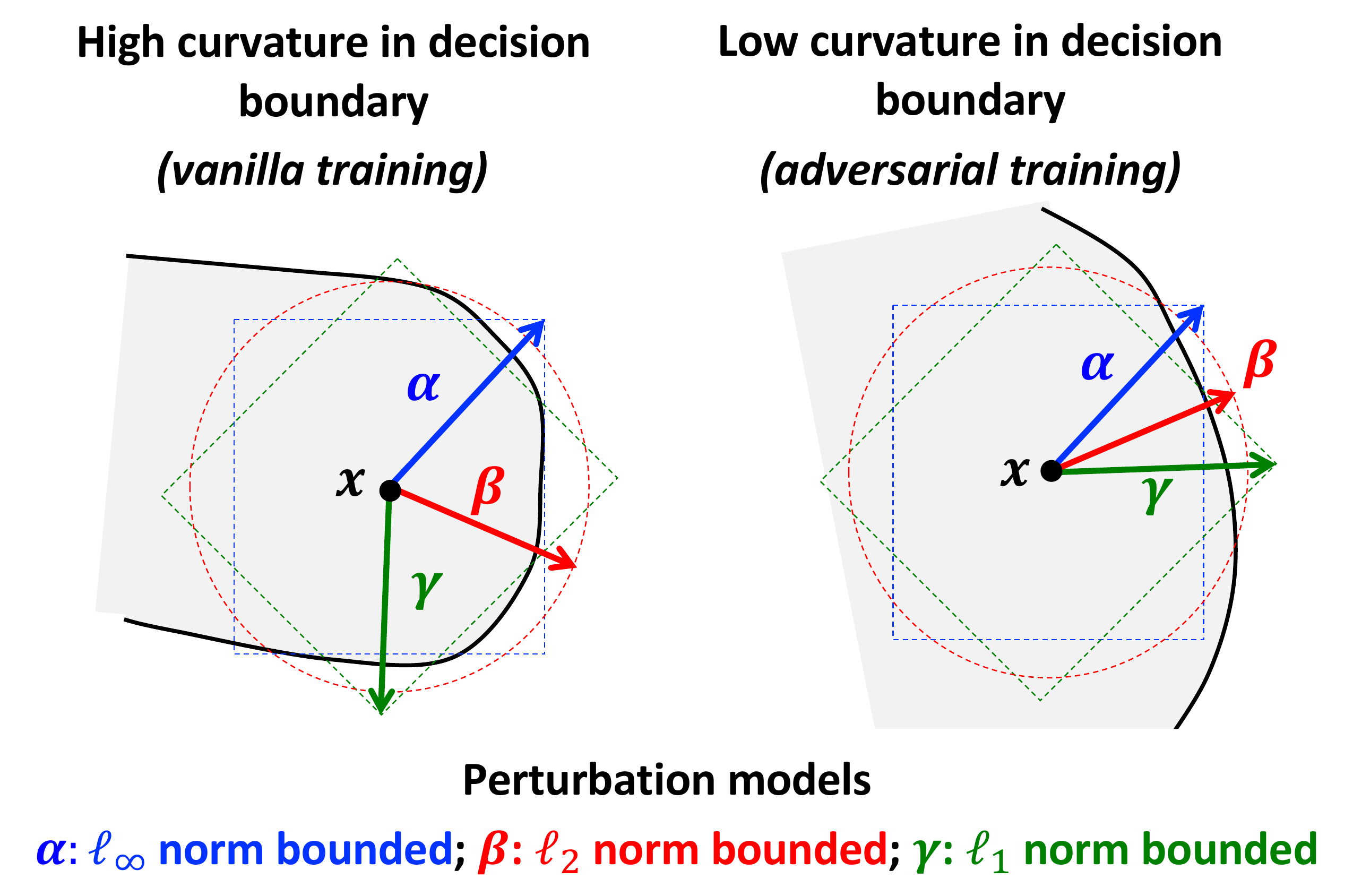}
	\end{center}
	\vspace{-0.4cm}
\caption{Illustration of the role of decision boundary curvature on the distinction between different types of perturbations $\bm{\alpha}$, $\bm{\beta}$ and $\bm{\gamma}$ of the given input $\vx$.}
	\label{fig::Intuition}
	\vspace{-0.5cm}
\end{wrapfigure}

In this section, we employ subspace methods to comprehend the distinction between $\ell_\infty$, $\ell_2$ and $\ell_1$ perturbations. %
For each input $\vx_i\in \mathbb{R}^D$ in dataset $X$, consider adversarial perturbations $\bm{\alpha}_i$, $\bm{\beta}_i$, and $\bm{\gamma}_i$ bounded within $\ell_\infty$, $\ell_2$, and $\ell_1$ norms, respectively.

We begin with a hypothesis (see  Fig.~\ref{fig::Intuition}): \emph{The perturbations $\bm{\alpha}$, $\bm{\beta}$, and $\bm{\gamma}$ corresponding to input $\vx$ have directions that differ significantly if the curvature of the decision boundary is high in the neighborhood of $\vx$. Conversely, if the curvature of the decision boundary is low, the perturbations $\bm{\alpha}$, $\bm{\beta}$, and $\bm{\gamma}$ tend to point in similar directions.} %

Since, prior works~\citep{dezfooli2018robustness,moosavi2019robustness} have found that single-attack AT reduces the curvature of the decision boundary, we test our hypothesis by studying the following two networks on CIFAR-10 data: a \emph{non-robust} ResNet18 $f^{\text{van}}_\theta$ trained using vanilla training, and a \emph{robust} ResNet18 $f^{\text{rob}}_\theta$ trained using the TRADES~\citep{zhang2019theoretically} AT framework employing $\ell_\infty$ perturbations.

\begin{figure}[t]
	\begin{center}
		\includegraphics[width=0.9\linewidth]{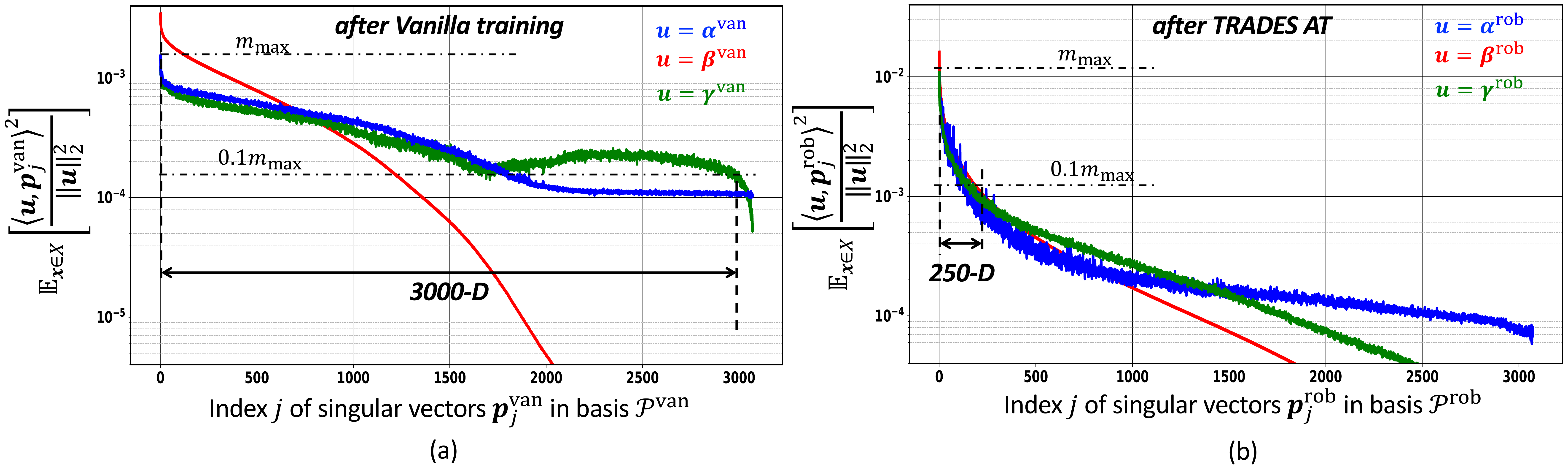}
	\end{center}
	\vspace{-0.4cm}
\caption{Normalized mean squared projections of three perturbation types on the singular vector basis $\mathcal{P}^{\kappa}$ of $\ell_2$ perturbations of ResNet18 on CIFAR-10 after: (a) vanilla training ($\kappa\equiv\text{van}$), and (b) TRADES training ($\kappa\equiv\text{rob}$). %
The singular vectors $\vp_i^{\kappa}$ comprising $\mathcal{P}^{\kappa}=\{\vp_1^{\kappa},\dots,\vp_D^{\kappa}\}$ are ordered in descending order of their singular values.}
	\label{fig::SubspaceObs}
	\vspace{-0.4cm}
\end{figure}

We compute perturbations $\bm{\alpha}_i$, $\bm{\beta}_i$, and $\bm{\gamma}_i$ for each $\vx_i\in X$ for both networks, \ie, $\kappa\in\{\text{van},\text{rob}\}$. We compute the singular vector basis  $\mathcal{P}^{\kappa}$ for the set of $\ell_2$ bounded perturbations $\Delta^{\kappa} =\{\bm{\beta}_1^{\kappa},\dots,\bm{\beta}_{|X|}^{\kappa}\}$. %
The normalized mean squared projections of the three types of perturbation vectors on the singular vector basis $\mathcal{P}^{\kappa}$ of vanilla trained ResNet-18  ($\mathcal{P}^{\text{van}}$)(Fig.~\ref{fig::SubspaceObs}(a)) and TRADES trained ResNet-18 ($\mathcal{P}^{\text{rob}}$)(Fig.~\ref{fig::SubspaceObs}(b)) shows a clear contrast.

The perturbations of a vanilla trained network roll-off gradually to occupy a larger subspace as indicated in Fig.~\ref{fig::SubspaceObs}(a). Specifically, the projections of $\bm{\alpha}$ and $\bm{\gamma}$ occupy almost all 3000 directions in the basis $\mathcal{P}^{\text{van}}$ since their mean squared projections are within $\sim 10\%$ of the maximum value $m_\text{max}$. This shows that the dominant singular vectors of $\bm{\beta}$ are not well-aligned with $\bm{\alpha}$ and $\bm{\gamma}$ in a vanilla trained network. With TRADES AT (Fig.~\ref{fig::SubspaceObs}(b)), however, all three types of perturbations are \emph{squeezed} into a much \emph{smaller} subspace spanning only the top 250 singular vectors in the perturbation basis $\mathcal{P}^{\text{rob}}$. Outside these 250 dimensions, the mean squared projections fall to $<10\%$ of their maximum value. 

In summary, the results in Fig.~\ref{fig::SubspaceObs} validate the hypothesis that single-attack AT increases the average alignment of different perturbation types due to the reduction in the decision boundary curvature. 
In Sec.~\ref{sec::technique}, we exploit this behavior of single-attack $\ell_\infty$ AT to improve its robustness against the union of multiple perturbation models via SNAP. %

%% file: ellinoise.tex
\section{Shaped Noise Augmented Processing (SNAP)}
\label{sec::technique}

\begin{wrapfigure}[]{r}{0.37\linewidth}
	\vspace{-1.1cm}
	\begin{center}
		\includegraphics[width=\linewidth]{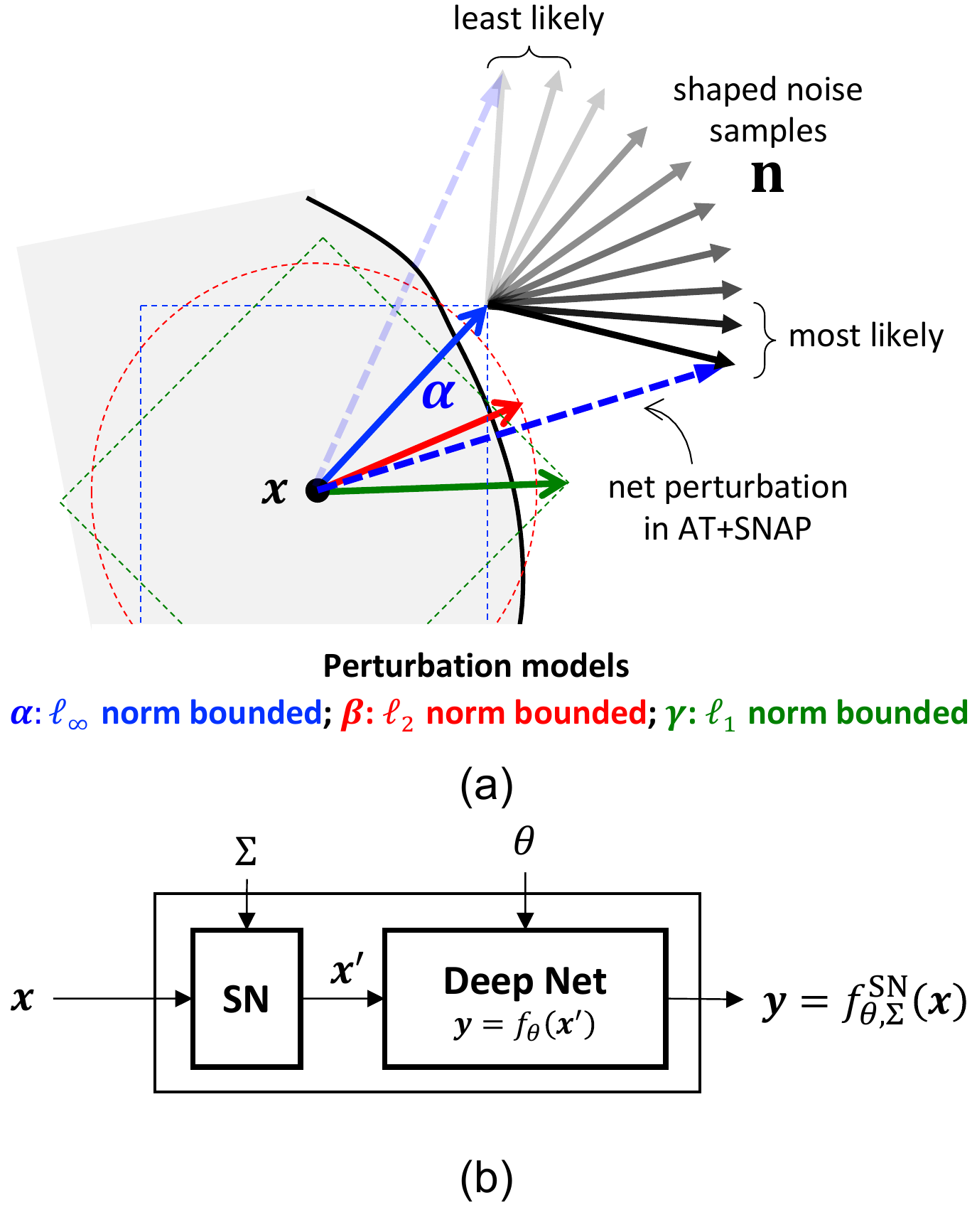}
	\end{center}
	\vspace{-0.5cm}
\caption{SNAP: (a) intuition underlying SNAP (not an exact depiction), and (b) SNAPnet $f^{\text{SN}}_{\theta, \Sigma}(\vx)$ constructed from a given deep net $f_\theta(\vx)$ by prepending a shaped noise (SN) augmentation layer which perturbs the primary input $\vx$ with noise $\rvn$ whose distribution parameter $\Sigma$ is learned during AT along with the base network parameter $\theta$.
}
	\label{fig::ElliNoiseConcept}
	\vspace{-0.8cm}
\end{wrapfigure}
We show that single-attack AT can be enhanced to address multiple perturbations by introducing noise to appropriately \emph{wiggle} the $\ell_\infty$-bounded perturbations (Fig.~\ref{fig::ElliNoiseConcept}(a)). 
However, to do so, the noise distribution needs to be \emph{chosen} and \emph{shaped} appropriately to minimize its impact on natural accuracy and robustness to $\ell_\infty$-bounded perturbations.

We experiment with both $\ell_\infty$ and $\ell_2$ perturbations in single-attack AT frameworks and find $\ell_\infty$-AT to be suitable for our proposed shaped noise augmentation (see Sec.~\ref{subsec::NoiseDistRes} for details). Hence, in this section, we describe SNAP for single-attack AT frameworks employing $\ell_\infty$ perturbations.

\subsection{SNAPnet}
A deep net $f_\theta( \vx) : \mathbb{R}^D \rightarrow \{0,1\}^C$ parametrized by $\theta$ maps the input $\vx\in \mathbb{R}^D$ to a one-hot vector $\vy\in\{0,1\}^C$ over $C$ classes. 

We construct a SNAP-based deep net (SNAPnet)  $f^{\text{SN}}_{\theta,\Sigma}( \vx )$ by introducing an additive shaped noise (SN) layer  (Fig.~\ref{fig::ElliNoiseConcept}(b)), where the noise distribution parameter $\Sigma$ is learned during training. Formally, 
\begin{align}
\vy = f^{\text{SN}}_{\theta, \Sigma}( \vx  ) = f_\theta\big(  \vx + \rvn \big) = f_\theta\big(  \vx + V\Sigma\rvn_0 \big), %
\label{eq::ElliNoiseModelDef}
\end{align}
where $ \rvn_0\sim \mathcal{L}(0,\mathbf{I}_{D\times D})$ is a zero-mean isotropic Laplace noise vector, $\Sigma =\text{Diag}[\sigma_1, \dots, \sigma_D]$ is a distribution parameter %
denoting 
its per-dimension standard deviation, $\mathbf{I}_{D\times D}$ denotes the $D\times D$ identity matrix, and $V=[\vv_1, \dots,  \vv_D]$ denotes a basis in $\mathbb{R}^D$. We also studied Gaussian and Uniform distributed $\rvn_0$, but empirically  find  the Laplace distribution to yield better results (Sec.~\ref{subsec::NoiseDistRes}). We use $V=\mathbf{I}_{D\times D}$ for all our experiments in the main text and study other options for $V$ in the Appendix. 

The final classification decision $d$ is computed via  
\begin{align}
    d = \arg\max_{c} \bigg[ \mathbb{E}_\rvn\big[ \vy \big] \bigg]_c,
    \label{eq::ElliNoiseDecRule} %
\end{align}
where $[\va]_c$ denotes the $c$-th element of vector $\va$. Note, the  shaped noise perturbs the input $\vx$ with a noise source $\rvn=V\Sigma\rvn_0$ (Eq.~(\ref{eq::ElliNoiseModelDef})). 
The distribution parameter $\Sigma$ is learned in the presence of any standard AT method \citep{madry2018towards,zhang2019theoretically,shafahi2019adversarial} used for learning deep net parameters $\theta$ as described next. %

\subsection{Training SNAPnet}
Algorithm~\ref{alg::algTRADESwSSN} summarizes the procedure for training SNAPnet $f_{\theta,\Sigma}^{\text{SN}}(\vx)$. 
In each epoch, an arbitrary AT method $\text{BASE}()$ (line $2$) updates network parameters $\theta$ with input perturbed by noise $\rvn$. Here
$\text{BASE}()$ can be any established AT framework \citep{madry2018towards,zhang2019theoretically,shafahi2019adversarial,wong2020fast} {employing $\ell_\infty$ perturbation}.
\begin{algorithm}[t]
	\small
	\caption{\small Training SNAPnet}
	\label{alg::algTRADESwSSN}
	\hspace*{\algorithmicindent} \textbf{Input:} training set $X$; basis $V=[\vv_1, \dots, \vv_D]$; total noise power $P_{\text{noise}}$; minibatch size $r$; baseline training method BASE; noise variance update frequency $U_f$; Total number of epochs $T$  \\ 
	\hspace*{\algorithmicindent} \textbf{Initialize:} noise variances $\Sigma_0=\text{Diag}[\sigma_{1,0}, \dots, \sigma_{D,0}]$. \\ %
	\hspace*{\algorithmicindent} \textbf{Output:} robust network $f^{\text{SN}}_{\theta,\Sigma}$, noise variances $\Sigma_T=\text{Diag}[\sigma^2_{1,T}, \dots, \sigma^2_{D,T}]$.
	\begin{algorithmic}[1]
		\For{epoch $t=1\ldots T$}
			\For {mini-batch $B = \{\vx_1,\dots, \vx_r\}$} \quad	$ \theta \leftarrow 
			\text{BASE}_{\ell_\infty}\bigg(f^{\text{SN}}_{\theta,\Sigma_t}\big(\{\vx_i\}_{i=1}^r \big),\theta\bigg)$ \Comment{  {\color{blue}\small   BASE() Training} } \EndFor
		    \If {$t \mod U_f = 0$} \Comment{ {\color{blue} SNAP Distribution Update once every $U_f$ epochs }}
		        \For {mini-batch $B = \{\vx_1,\dots, \vx_r\}$ }
    				\State  { $ \{ \vx^{\text{adv}}_i \}_{i=1}^r \leftarrow 
    			    \text{PGD}^{(K)}_{\ell_2}\bigg(f^{\text{SN}}_{\theta,\Sigma_t}\big(\{\vx_i\}_{i=1}^r \big)\bigg)$; \quad $ \bm{\eta}_i = \vx^{\text{adv}}_i - \vx_i \hspace{0.2cm} \forall \hspace{0.2cm} i\in\{1,\dots,r\} $}
        			\State $\gamma_{j} \leftarrow \gamma_{j} + \sum_{i=1}^{r} \big(\langle \vv_j,\bm{\eta}_i \rangle\big)^2 \quad \forall j\in\{1,\dots,D\} $  \Comment{{\small\color{blue} Accumulate projections; See Eq.~(\ref{eqn::VarAlloc}})}
			    \EndFor
			    \State $ \sigma^2_{j,t+1} =  P_{\text{noise}}\frac{\sqrt{\gamma_{j}}}{\sum_{k=1}^D \sqrt{\gamma_{k}} }  \quad \forall  j\in\{1,\dots,D\}$  \Comment{{\color{blue}\small Normalize accumulated projections; See Eq.~(\ref{eqn::VarAlloc}})}
			\Else
			    \State{$\Sigma_{t+1} \leftarrow \Sigma_t$}
    		\EndIf
		\EndFor

	\end{algorithmic}
	
\end{algorithm}

The SNAP parameter $\Sigma$ is updated once every $U_f=10$ epochs via a \emph{SNAP distribution update}  (lines $4$-$10$). In this update, the per-dimension noise variance $\sigma^2_j$ is updated proportional to the root mean squared projection of the adversarial perturbations $\bm{\eta}$ %
on the basis $V$ %
given a total noise constraint $\sum_{j=1}^D \sigma^2_{j}=P_\text{noise}$, where $P_\text{noise}$ denotes the total noise power. Formally, 
\vspace{-0.3cm}
\begin{align}
    \displaystyle
    \sigma^2_{j} \propto \sqrt{\mathbb{E}_{\vx\in X}\big(\langle \bm{\eta}, \vv_j \rangle^2\big)}\quad \text{s.t.}\quad \sum_{j=1}^D \sigma^2_{j}=P_\text{noise}, 
    \label{eqn::VarAlloc}
\end{align}
where $\bm{\eta}$ is the $\ell_2$ norm-bounded PGD adversarial perturbation for the given input $\vx \in X$ (line 6).  Note that these $\ell_2$ perturbations are employed \emph{only} for noise shaping and are distinct from the $\ell_\infty$ perturbations employed by BASE() AT (line 2). Also, $\ell_\infty$ perturbations cannot be used here since their projections are constant $\forall  j$ when $V=\mathbf{I}_{D\times D}$, whereas employing $\ell_1$ perturbations leads to poor shaping due to high sparsity. %

Thus, in SNAP, the average squared $\ell_2$ norm of the noise vector $\rvn$ is held constant at $P_\text{noise}$ while adapting the noise variances in the individual dimensions so as to align the noise vectors with the adversarial perturbations \emph{on average}. Intuitively, the decision boundary is pushed aggressively in those directions. %

\subsection{Remarks}

Note that the SNAP distribution update  is distinct from BASE() AT. Hence, SNAP doesn't require any hyperparameter tuning in BASE(). For fairness to baselines we keep all hyperparameters identical when introducing SNAP in all our experiments. However, SNAP  introduces a new hyperparameter $P_\text{noise}$, which permits to trade adversarial robustness $\mathcal{A}^{(U)}_\text{adv}$ for natural accuracy $\mathcal{A}_\text{nat}$. This trade-off is explored  in Sec.~\ref{subsec::NoisePowerRes}.

The computational overhead of SNAP is small ($\sim 10\%$) since the \emph{SNAP Distribution Update} occurs once in 10 epochs using just 20\% of the training data to update the noise standard deviations $\sigma_j$. We provide more details about the \emph{SNAP Distribution Update}  in the Appendix.

%% file: results.tex
\section{Experimental Results}
\label{sec::Results}
\subsection{Setup}
Following  experimental settings of prior work~\citep{zhang2019theoretically,shafahi2019adversarial,maini2019adversarial}, we employ a ResNet-18 network for CIFAR-10 experiments and both ResNet-50 and ResNet-101 networks for ImageNet experiments. Accuracy on clean test data is referred to with $\mathcal{A}_\text{nat}$ and accuracy on adversarially perturbed test data is referred to via $\mathcal{A}^{(\ell_\infty)}_\text{adv}$, $\mathcal{A}^{(\ell_2)}_\text{adv}$, and $\mathcal{A}^{(\ell_1)}_\text{adv}$, for $\ell_\infty$, $\ell_2$, and $\ell_1$ norm bounded perturbations, respectively. Accuracy against the \emph{union} of all three perturbations is denoted by $\mathcal{A}^{(U)}_\text{adv}$.

For a fair robustness comparison, our evaluation setup closely follows the setup of~\citet{maini2019adversarial} for CIFAR-10 data: (1) choose norm bounds $\epsilon=(0.031,0.5,12.0)$ for ($\ell_\infty$, $\ell_2$, $\ell_1$) perturbations, respectively; (2) scale norm bounds for images to lie between $[0,1]$; (3) choose the PGD attack configuration to be \emph{100 iterations with 10 random restarts} for all perturbation types%
\footnote{Following~\citet{maini2019adversarial}, we also run all attacks on a subset of the first 1000 test examples with 10 random restarts for CIFAR-10 data.}; and (4) estimate $\mathcal{A}^{(U)}_\text{adv}$ as the fraction of test data that is \emph{simultaneously} resistant to all three perturbation models. %

Following the guidelines of \citet{tramer2020adaptive}, we carefully design \emph{adaptive} PGD attacks that target the full defense -- SN layer -- since SNAPnet is end-to-end differentiable. Specifically, we backpropagate to primary input $\vx$ through the SN layer (see Fig.~\ref{fig::ElliNoiseConcept}). Thus, the final shaped noise distribution is exposed to the adversary. We also account for the expectation $\mathbb{E}_\rvn[\cdot]$ in Eq.~(\ref{eq::ElliNoiseDecRule}) by explicitly averaging deep net logits over $N_0(=8)$ noise samples \emph{before} computing the gradient, which eliminates any gradient obfuscation, and is known to be the strongest attack against noise augmented models~\cite{salman2019provably}. In the Appendix~\ref{app::StressTest} we also show robustness stress tests and evaluate more attacks. 

On CIFAR-10 data, we compare with the following seven key SOTA AT frameworks: PGD~\cite{madry2018towards}, TRADES~\cite{zhang2019theoretically}, FreeAdv~\cite{shafahi2019adversarial}, FastAdv~\cite{wong2020fast}, AVG~\cite{tramer2019adversarial}, MSD~\cite{maini2019adversarial},  PAT~\cite{laidlaw2021perceptual}. We also compare with two randomized smoothing  frameworks~\cite{cohen2019certified,salman2019provably} in the Appendix~\ref{app::RandSmooth}. Thanks to their GitHub code releases, we first successfully reproduce their results with a ResNet-18 network in our environment. In the case of PAT~\citep{laidlaw2021perceptual}, we evaluate and compare with their pretrained ResNet-50 model on CIFAR-10. We compare all training times on a single NVIDIA P100 GPU. On ImageNet data, we primarily compare to FreeAdv~\cite{shafahi2019adversarial}. We train ResNet-50 and its SNAPnet version with FreeAdv on a Google Cloud server with four NVIDIA P100 GPUs to compare their accuracy and training times. We provide all hyperparameters in Appendix~\ref{app::Details}. Our code and pretrained models are available at \href{https://github.com/adpatil2/SNAP}{https://github.com/adpatil2/SNAP}.

\vspace{-0.2cm}
\subsection{Ablation Studies}
\begin{table}[!t]
\begin{minipage}{\textwidth}
\begin{minipage}[b]{0.49\textwidth}
  \centering
	\resizebox{\columnwidth}{!}{
	\begin{tabular}{|c|>{\centering\arraybackslash}p{0.9cm}|>{\centering\arraybackslash}p{1cm}>{\centering\arraybackslash}p{0.9cm}>{\centering\arraybackslash}p{0.9cm}|>{\centering\arraybackslash}p{0.9cm}| } %
		\hline
		{\makecell{Method}} &  {\makecell{$\mathcal{A}_\text{nat}$} } & {\makecell{$\mathcal{A}^{(\ell_\infty)}_\text{adv}$ \\ {\scriptsize $\epsilon=0.03$} }} & {\makecell{$\mathcal{A}^{(\ell_2)}_\text{adv}$\\ {\scriptsize $\epsilon=0.5$} }} &  {\makecell{$\mathcal{A}^{(\ell_1)}_\text{adv}$\\ {\scriptsize $\epsilon=12$}  }} & {\makecell{$\mathcal{A}^{(U)}_\text{adv}$} } \\ [1.2ex]
		\hline
		\hline
		\multicolumn{6}{|c|}{\textbf{PGD AT with $\ell_\infty$ perturbations }} \\
		\hline
		\hline
		{\makecell{PGD}}  & {84.6} & \textbf{48.8} & 62.3 & {15.0} & {15.0}   \\
    		 \textbf{+SNAP[G]}  & {80.7}  & {45.7}  & {66.9}  &  {34.6} &  {31.9} \\
    		 \textbf{+SNAP[U]}  & \textbf{85.1}  & {42.7}  & \textbf{66.7}  &  {28.6} &  {26.6}  \\
    		\textbf{+SNAP[L]}  & {83.0}  & {44.8}  & \textbf{68.6}  &  \textbf{40.1} &  \textbf{35.6} \\
    	\hline
		\hline
		\multicolumn{6}{|c|}{\textbf{PGD AT with $\ell_2$ perturbations }} \\
		\hline
		\hline
		{\makecell{PGD}}  & \textbf{89.3} & {28.8} & \textbf{67.3} & {31.8} & {25.1}   \\
    		 \textbf{+SNAP[G]}  & {83.0}  & \textbf{35.0}  & {65.8}  &  {39.9} &  {30.2} \\
    		 \textbf{+SNAP[U]}  & {86.4}  & {32.3}  & {66.7}  &  {30.2} &  {25.0}  \\
    		\textbf{+SNAP[L]}  & {84.8}  & {33.4}  & {66.1}  &  \textbf{42.5} &  \textbf{30.8} \\
    	\hline %
	\end{tabular}}
	\vspace{0.1cm}
	\captionof{table}{ResNet-18 CIFAR-10 results showing the impact of SNAP augmentation of   PGD~\citep{madry2018towards} AT framework with $\ell_\infty$ (\emph{top}) and $\ell_2$ (\emph{bottom}) perturbations where [G], [U], and [L], denote shaped Gaussian, Uniform, and Laplace noise.}
	\label{tab::NoiseDistComp}
\end{minipage}
\hfill
\begin{minipage}[b]{0.49\textwidth}
 \centering
 \resizebox{\columnwidth}{!}{
 \begin{tabular}{ |c|>{\centering\arraybackslash}p{0.9cm}|>{\centering\arraybackslash}p{1cm}>{\centering\arraybackslash}p{0.9cm}>{\centering\arraybackslash}p{0.9cm}|>{\centering\arraybackslash}p{0.9cm} |  } %
		\hline
		{\makecell{Method}} &  {\makecell{$\mathcal{A}_\text{nat}$} } & {\makecell{$\mathcal{A}^{(\ell_\infty)}_\text{adv}$ \\ {\scriptsize $\epsilon=0.03$} }} & {\makecell{$\mathcal{A}^{(\ell_2)}_\text{adv}$\\ {\scriptsize $\epsilon=0.5$} }} &  {\makecell{$\mathcal{A}^{(\ell_1)}_\text{adv}$\\ {\scriptsize $\epsilon=12$}  }} & {\makecell{$\mathcal{A}^{(U)}_\text{adv}$} } \\ [1.2ex]
		\hline
		\hline
		\multicolumn{6}{|c|}{\textbf{High Complexity AT with $\ell_\infty$ perturbations}} \\
		\hline
		\hline
		\makecell{PGD}  & \textbf{84.6} & \textbf{48.8} & 62.3 & {15.0} & {15.0}   \\
    		 \textbf{+SNAP} &    {83.0}  &  {44.8}  & \textbf{68.6}  &  \textbf{40.1} &  \textbf{35.6}  \\
    	\hline %
		\hline %
		\makecell{TRADES}  & \textbf{82.1} & \textbf{50.2} & 59.6 & {19.8} & {19.7}  \\
    		 \textbf{+SNAP}  & {80.9}  & {45.2}  & \textbf{66.9}  &  \textbf{46.6} &  \textbf{41.2}  \\
    		
    	\hline
		\hline
		\multicolumn{6}{|c|}{\textbf{Low Complexity AT with $\ell_\infty$ perturbations}} \\
		\hline
		\hline
		\makecell{FreeAdv}   & {81.7} & \textbf{46.1} & 59 & {15.0} & {15.0} \\
    		 \textbf{+SNAP} &  \textbf{83.5}  & {39.7}  & {\textbf{66.2}}  & \textbf{34.3} & \textbf{29.6}  \\
		\hline
		\hline %
		\makecell{FastAdv}   & \textbf{85.7} & \textbf{46.2} & 60.0 & {13.2} & {13.2} \\
    		 \textbf{+SNAP} &  {84.2}  & {40.4}  & {\textbf{67.9}}  & \textbf{36.6} & \textbf{30.8}  \\
		\hline
	\end{tabular}}
	\vspace{0.1cm}
 \captionof{table}{ResNet-18 CIFAR-10 results showing the impact of SNAP augmentation of established $\ell_\infty$-AT frameworks. %
 The computational overhead of SNAP is limited to $\sim10\%$.}
 \label{tab::AdvTrainingComp}
\end{minipage}
\end{minipage}
\vspace{-0.8cm}
\end{table}

\subsubsection{Impact of Noise Distribution and Model of BASE() AT Perturbations}
\label{subsec::NoiseDistRes}
In this subsection, we first study the impact of employing $\ell_\infty$ \vs $\ell_2$ perturbations in BASE AT() (see line 2 in Alg.~\ref{alg::algTRADESwSSN}) on $\mathcal{A}^{(U)}_\text{adv}$. For each choice, we further experiment with three distributions for the SN layer in Fig.~\ref{fig::ElliNoiseConcept}(b) viz. Gaussian, Uniform, and Laplace. We don't consider $\ell_1$ perturbations in BASE AT() since \citet{maini2019adversarial} showed that employing $\ell_1$ single-attack AT  achieves very low robustness to all attacks. %
We choose PGD~\cite{madry2018towards} AT as BASE AT() for this ablation study. For a fair comparison across the noise distributions, we fix $P_\text{noise}=160$, enforcing all noise vectors to have the same average $\ell_2$ norm. For each distribution, the noise is shaped per the procedure summarized in Alg.~\ref{alg::algTRADESwSSN}. 

As observed in Table~\ref{tab::NoiseDistComp}, $\ell_\infty$-PGD AT achieves much lower $\mathcal{A}^{(U)}_\text{adv}$ than $\ell_2$-PGD AT, an observation also reported by \citet{maini2019adversarial}. With SNAP, however, we find that there is an interaction between the perturbation model in PGD AT and the noise distribution in SNAP. For instance, SNAP[U] enhances $\mathcal{A}^{(U)}_\text{adv}$ by 11\% with $\ell_\infty$-PGD AT while not achieving any improvement with $\ell_2$-PGD AT. In fact, SNAP appears to be particularly suitable for $\ell_\infty$-AT, since it always improves $\mathcal{A}^{(U)}_\text{adv}$ by 11\%-to-20.6\% irrespective of the noise distribution.

Finally, of the three noise distributions, we find the Laplace distribution to be distinctly superior, achieving the highest $\mathcal{A}^{(U)}_\text{adv}$ (35.6\% and 30.8\%) due to a significant improvement in $\mathcal{A}^{(\ell_1)}_\text{adv}$ for both $\ell_\infty$ and $\ell_2$ PGD AT, respectively. The superiority of the Laplace distribution in achieving high $\mathcal{A}^{(\ell_1)}_\text{adv}$ stems from its heavier tail compared to the Gaussian and Uniform distributions with the same variance. Shaped Laplace noise generates the highest fraction of extreme values in a given noise sample. Hence, it is more effective in improving accuracy against $\ell_1$-bounded attacks, which are the strongest when perturbing few pixels by a large magnitude~\cite{maini2019adversarial, tramer2019adversarial}. We discuss this further in the Appendix~\ref{app::NoiseHist}. 
Henceforth, unless otherwise mentioned, we choose Laplace  noise for SNAP and $\ell_\infty$ perturbations for BASE() AT as the default setting since it achieves the highest $\mathcal{A}^{(U)}_\text{adv}$. %

\subsubsection{Impact of $P_\text{noise}$}
\label{subsec::NoisePowerRes}
\begin{wrapfigure}[]{r}{0.45\linewidth}
    \vspace{-0.7cm}
	\begin{center}
		\includegraphics[width=\linewidth]{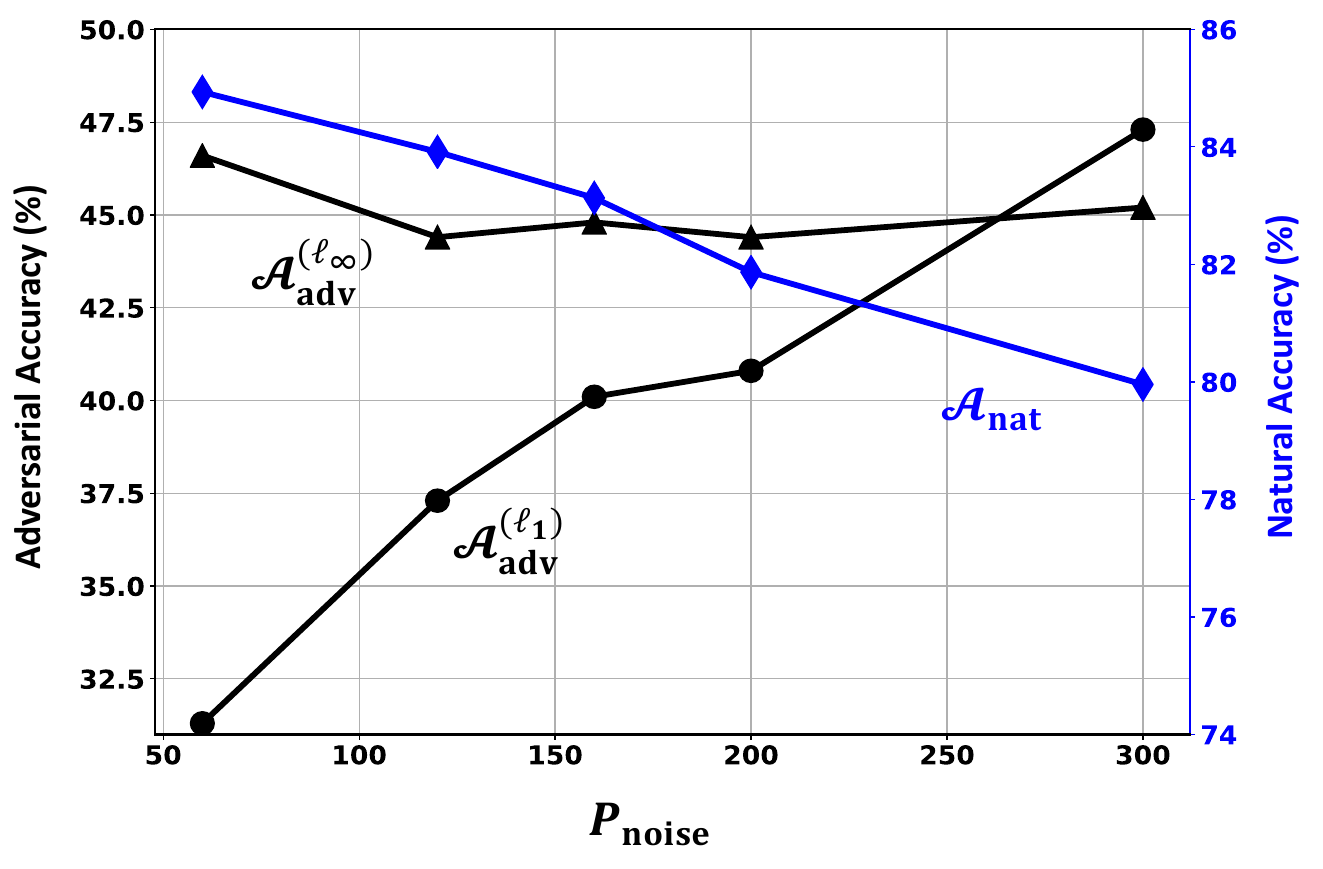}
	\end{center}
	\vspace{-0.4cm}
	\caption{ResNet-18 CIFAR-10 results: adversarial accuracy $\mathcal{A}^{(\ell_1)}_\text{adv}$, $\mathcal{A}^{(\ell_\infty)}_\text{adv}$, and natural accuracy $\mathcal{A}_\text{nat}$ \vs total noise power $P_\text{noise}$ for PGD+SNAP.}
	\label{fig::PnoiseRole} 
	\vspace{-1.2cm}
\end{wrapfigure}
Next, we explore the impact of the SNAP hyperparameter $P_\text{noise}$, which constrains the average squared $\ell_2$ norm of the noise vector $\rvn$. It enables to trade between adversarial and natural accuracy. %

Fig.~\ref{fig::PnoiseRole} shows that, as $P_\text{noise}$ increases, $\mathcal{A}^{(\ell_1)}_\text{adv}$ improves from 31\% to 47\%,  accompanied by a graceful ($5\%$) drop in $\mathcal{A}_\text{nat}$ and a small drop of 2\% in $\mathcal{A}^{(\ell_\infty)}_\text{adv}$ that stabilizes to $\approx 45\%$. These results show: (1) SNAP preserves the impact of $\ell_\infty$ perturbations which is not surprising since PGD AT~\cite{madry2018towards} explicitly includes those, and (2) $P_\text{noise}$ provides an explicit knob to control the $\mathcal{A}_\text{nat}$ \vs $\mathcal{A}_\text{adv}$  trade-off.  Henceforth, we choose $P_\text{noise}$ values that incur $< 1.5\%$ drop in $\mathcal{A}_\text{nat}$ for all SNAP+AT experiments.

\subsubsection{SNAP augmented SOTA AT Frameworks}
\label{subsubsec::ATaug}

Table~\ref{tab::AdvTrainingComp} shows the effectiveness of SNAP for four SOTA AT frameworks: high complexity frameworks, such as  PGD~\citep{madry2018towards}, TRADES~\cite{zhang2019theoretically}, and low complexity frameworks such as FreeAdv~\cite{shafahi2019adversarial}, FastAdv~\cite{wong2020fast}. All are trained against $\ell_\infty$ attacks with $\epsilon=0.031$. As expected, while they achieve high $\mathcal{A}^{(\ell_\infty)}_\text{adv}$, their  $\mathcal{A}^{(\ell_2)}_\text{adv}$ and $\mathcal{A}^{(\ell_1)}_\text{adv}$ are lower.

For high-complexity AT, SNAP enhances $\mathcal{A}^{(\ell_2)}_\text{adv}$ and $\mathcal{A}^{(\ell_1)}_\text{adv}$ by $\sim6\%$ and $\sim25\%$, respectively, while incurring only a drop of $\sim 5\%$ in $\mathcal{A}^{(\ell_\infty)}_\text{adv}$. Thus overall, SNAP  improves robustness ($\mathcal{A}^{(U)}_\text{adv}$) by $\sim 20\%$ against the \emph{union} of the three perturbation models. Note that this robustness improvement comes at only a $\sim 1\%$ drop in $\mathcal{A}_\text{nat}$ (see Table~\ref{tab::AdvTrainingComp}). For low-complexity ATs, SNAP improvements in union robustness ($\mathcal{A}^{(U)}_\text{adv}$) are also significant ($\sim 15\%$). Again, presence of SNAP  improves $\mathcal{A}^{(\ell_2)}_\text{adv}$ and $\mathcal{A}^{(\ell_1)}_\text{adv}$. This time the drop in $\mathcal{A}^{(\ell_\infty)}_\text{adv}$ is $\sim7\%$. We believe this is due to the fact that these frameworks employ weaker single-step attacks during  training. Note that in the case of FreeAdv+SNAP, we actually observe a $\sim 2\%$ \emph{increase} in $\mathcal{A}_\text{nat}$, a trend we also observe in the ImageNet experiments described later.

\begin{table*}[t]
	\centering
	\small
	\resizebox{0.75\linewidth}{!}{
    \begin{tabular}{ |c|>{\centering\arraybackslash}p{1.8cm}|>{\centering\arraybackslash}p{1.2cm}|>{\centering\arraybackslash}p{0.9cm}|>{\centering\arraybackslash}p{0.9cm}|>{\centering\arraybackslash}p{1.8cm}| } %
	\hline
	{\makecell{Method}} & \makecell{LR schedule} & Epochs & {\makecell{$\mathcal{A}_\text{nat}$} } & {\makecell{$\mathcal{A}^{(U)}_\text{adv}$} }  & Total time (minutes) \\ [1.2ex]
	\hline
	\hline
	\multicolumn{6}{|c|}{\textbf{Set A: Total Time $\geq$ 12 Hrs}} \\
	\hline
	\hline
	\makecell{AVG 50 Step  \cite{tramer2019adversarial} } & \makecell{cyclic}  & 50 &  {84.8}  & {40.4}  & 4217  \\
	\makecell{AVG 20 Step  \cite{tramer2019adversarial} } & \makecell{cyclic}  & 50 &  \textbf{85.6}  & {40.4}  & 1834  \\
	\makecell{AVG 10 Step  \cite{tramer2019adversarial} } & \makecell{cyclic}  & 50 &  \textbf{86.7}  & {38.9}  & \textbf{956}  \\
	\makecell{PAT  \cite{laidlaw2021perceptual} } & \makecell{step}  & 100 &  {82.4}  & {36.6}  & 1364  \\
    \makecell{MSD 50 Step  \cite{maini2019adversarial} } & \makecell{cyclic}  & 50 &  {81.7}  & \textbf{47.0}  & 1693  \\
    \makecell{MSD 30 Step  \cite{maini2019adversarial} } & \makecell{cyclic} & 50 &  {82.4}  & \textbf{44.9}  & \textbf{978}  \\
    \hline
	\hline
	\multicolumn{6}{|c|}{\textbf{Set B: 8 Hrs $<$ Total Time $<$ 12 Hrs}} \\
	\hline
	\hline
	\makecell{AVG 5 Step  \cite{tramer2019adversarial} } & \makecell{cyclic} & 50 &  \textbf{87.8}  & {33.7}   & {489}  \\
	\makecell{MSD 20 Step  \cite{maini2019adversarial} } & \makecell{cyclic} & 50 &  {83.0}  & {37.3}   & {690}  \\
	\makecell{TRADES \cite{zhang2019theoretically} } & \makecell{step} & 100 &  {82.0}  & {19.7}   & \textbf{516}  \\
	\makecell{ \textbf{TRADES+SNAP} } & \makecell{step} & 100 &  {80.9}  & \textbf{41.2}   & 566  \\
	\hline
	\hline
	\multicolumn{6}{|c|}{\textbf{Set C: 5 Hrs $<$ Total Time $<$ 8 Hrs}} \\
	\hline
	\hline
    \makecell{MSD 10 Step  \cite{maini2019adversarial} } & \makecell{cyclic} & 50 &  {83.6}  & {33.3}   & \textbf{342}  \\
    \makecell{PGD \cite{madry2018towards} } & \makecell{step} & 100 &  \textbf{84.6}  & {15.0}   & {354}  \\
    \makecell{ \textbf{PGD+SNAP} } & \makecell{step} & 100 &  {83.0}  & \textbf{35.6}   & {403}  \\
    \hline
	\hline
	\multicolumn{6}{|c|}{\textbf{Set D: 2 Hrs $<$ Total Time $<$ 5 Hrs}} \\
	\hline
	\hline
	\makecell{AVG 2 Step  \cite{tramer2019adversarial} } & \makecell{cyclic} & 50 &  \textbf{88.4}  & {22.0}   & {232}  \\
	\makecell{MSD 5 Step  \cite{maini2019adversarial} } & \makecell{cyclic} & 50 &  \textbf{84.0}  & {12.6}   & \textbf{185}  \\
    \makecell{PGD \cite{madry2018towards} } & \makecell{cyclic} & 50 &  {82.8}  & {15.7}   & \textbf{177}  \\
    \makecell{TRADES \cite{zhang2019theoretically} } & \makecell{cyclic} & 50 &  {80.0}  & {21.4}   & 258  \\
    \makecell{ \textbf{PGD+SNAP } } & \makecell{cyclic} & 50 &  {82.3}  & \textbf{33.5}   & {199}  \\
    \makecell{\textbf{TRADES+SNAP} } & \makecell{cyclic} & 50 &  {78.8}  & \textbf{40.8}   & 280  \\
    \hline
    \hline
	\multicolumn{6}{|c|}{\textbf{Set E: Total Time $<$ 2 Hrs}} \\
	\hline
	\hline
    \makecell{FreeAdv \cite{shafahi2019adversarial} } & \makecell{step} & 200 &  {81.7}  & {15.0}   & \textbf{66}  \\
    \makecell{FastAdv \cite{wong2020fast} } & \makecell{cyclic}  &  50  & \textbf{85.7}  & 13.2  &  \textbf{47}  \\
    \makecell{ \textbf{FreeAdv+SNAP} } & \makecell{step} & 200  &  {83.5}  & \textbf{29.6}   & 88  \\
    \makecell{ \textbf{FastAdv+SNAP} } & \makecell{cyclic} & 50  &  \textbf{84.2}  & \textbf{30.8}   & {69}  \\
    \hline
    \end{tabular}}
    \caption{CIFAR-10 results for comparing adversarial accuracy $\mathcal{A}^{(U)}_\text{adv}$ \vs training time (on single NVIDIA P100 GPU) for different AT frameworks and the improvements by introducing proposed SNAP technique. All frameworks except PAT~\cite{laidlaw2021perceptual} (which employs ResNet-50) employ ResNet-18.}
	\label{tab::CIFARTimeComp}
\end{table*}

\subsection{Robustness \vs Training Complexity}

Next we quantify adversarial robustness \vs training time trade-offs. %
Table~\ref{tab::CIFARTimeComp}  shows that SNAP augmentation of single-attack AT frameworks achieves the highest $\mathcal{A}^{(U)}_\text{adv}$, when training time is constrained to 12 hours (sets \textbf{B}, \textbf{C}, \textbf{D}, and \textbf{E}).

For instance, TRADES+SNAP achieves a 4\% higher $\mathcal{A}^{(U)}_\text{adv}(=41\%)$ than MSD-$20$ with 2 hours \emph{lower} training time (Set \textbf{B} in Table~\ref{tab::CIFARTimeComp}). Similarly, PGD+SNAP achieves a 2\% higher $\mathcal{A}^{(U)}_\text{adv}$ than MSD-$10$ while having a similar training time (Set \textbf{C}). Note that both PGD and TRADES here use 100 training epochs with standard step learning rate (LR) schedule, while MSD frameworks employ a cyclic learning rate  schedule to achieve superconvergence in 50 epochs.

\begin{table*}[t]
	\centering
	\resizebox{0.85\linewidth}{!}{
    \begin{tabular}{ |c|>{\centering\arraybackslash}p{1.5cm}|>{\centering\arraybackslash}p{1.5cm}>{\centering\arraybackslash}p{1.5cm}>{\centering\arraybackslash}p{1.5cm}|>{\centering\arraybackslash}p{1.5cm}|>{\centering\arraybackslash}p{1.5cm}| } %
	\hline
	{\makecell{Training}}  & {{$\mathcal{A}_\text{nat}$ (\%)} } & {\makecell{$\mathcal{A}^{(\ell_\infty)}_\text{adv}$ \\ {\scriptsize $\epsilon=2/255$} }} & {\makecell{$\mathcal{A}^{(\ell_2)}_\text{adv}$\\ {\scriptsize $\epsilon=2.0$ } }} &  {\makecell{$\mathcal{A}^{(\ell_1)}_\text{adv}$\\ {\scriptsize $\epsilon=72.0$} }} & {{$\mathcal{A}^{(U)}_\text{adv}$} } & \makecell{Total time \\ (minutes)} \\
	\hline
	\hline
	\multicolumn{7}{|c|}{\textbf{ResNet-50}} \\
    \hline
    \makecell{ FreeAdv {\small\citep{shafahi2019adversarial}}\\} & 61.7 & \textbf{47.8} & 19.9 & {14.8} & 12.6 &  \textbf{3590}  \\
    \hline
    \makecell{ \textbf{FreeAdv+SNAP}\\ } &   \textbf{ 66.8 } &   { 46.1 } &  \textbf{ 37.8 } &  \textbf{ 37.4 } & \textbf{32.4}   &  {3756} \\
    \hline
	\hline
	\multicolumn{7}{|c|}{\textbf{ResNet-101}} \\
    \hline
    \makecell{ FreeAdv {\small\citep{shafahi2019adversarial}}\\} & 65.4 & \textbf{51.8} & 22.8 & {18.8} & 16.1 &  \textbf{5678}  \\
    \hline
    \makecell{ \textbf{FreeAdv+SNAP}\\ } &   \textbf{ 69.7 } &   { 50.3 } &  \textbf{ 41.1 } &  \textbf{ 40.2 } & \textbf{35.4}   &  {5904} \\
    \hline
    \end{tabular}}
    \caption{ImageNet results: Iso-hyperparameter introduction of SNAP yields $\sim 20\%$ improvement in adversarial accuracy ($\mathcal{A}^{(U)}_\text{adv}$) with modest impact on training time for ResNet-50 and ResNet-101.}
	\label{tab::ImageNetComp}
	\vspace{-0.3cm}
\end{table*}

In Set \textbf{D}, \emph{following}  \citet{maini2019adversarial}, we employ a cyclic learning rate schedule  for PGD, TRADES, as well as for PGD+SNAP and TRADES+SNAP to achieve convergence in 50 epochs. %
Improvements in $\mathcal{A}^{(U)}_\text{adv}$ for PGD+SNAP and TRADES+SNAP are similar to those in Sets \textbf{B} and \textbf{C}. Most notably, PGD+SNAP with cyclic learning rate achieves $\sim 20\%$ and  11.5\% \emph{higher} $\mathcal{A}^{(U)}_\text{adv}$ than MSD-5 and AVG-2, respectively, while having a similar training time ($\sim 3$ hours). Set \textbf{E} augments the  data from  Table~\ref{tab::AdvTrainingComp} with  training times. FastAdv+SNAP and FreeAdv+SNAP achieve a high $\mathcal{A}^{(U)}_\text{adv}\sim 30\%$, while preserving the training efficiency of both FastAdv and FreeAdv. Notably, FastAdv+SNAP achieves $18\%$ higher $\mathcal{A}^{(U)}_\text{adv}$ than MSD-5, while being $\sim2.7\times$ more efficient to train.

\subsection{ImageNet Results}
\label{subsec::ImageNet}
Thanks to SNAP's low computational overhead combined with FreeAdv's fast training time, we are for the first time able to report adversarial accuracy of ResNet-50 and ResNet-101   against the union of $(\ell_\infty,\ell_{2},\ell_1)$ attacks on ImageNet.

We closely follow the evaluation setup of \citet{shafahi2019adversarial}. Specifically, we use 100 step PGD attack, one of the strongest adversaries considered by \citet{shafahi2019adversarial}, and evaluate on the entire test set. We first reproduce FreeAdv~\cite{shafahi2019adversarial} results using the \emph{same} hyperparameters and then introduce SNAP. All hyperparameter details are specified in the Appendix.

In order to clearly demonstrate the contrast between robustness to different perturbation models, we evaluate with $\epsilon=(2/255,2.0,72.0)$ for $(\ell_\infty,\ell_2,\ell_1)$ attacks, respectively.\footnote{Note that $\ell_2$ and $\ell_1$ norms of PGD perturbation with $\ell_\infty$ norm of $2/255$ can be as large as $\sim3.0$ and $\sim1100$ for images of size $224\times224\times3$.} As shown in Table~\ref{tab::ImageNetComp}, FreeAdv achieves a  high $\mathcal{A}^{(\ell_\infty)}_\text{adv}=47.8\%$ with ResNet-50, but a lower $\mathcal{A}^{(\ell_2)}_\text{adv}=20\%$ and $\mathcal{A}^{(\ell_1)}_\text{adv}=15\%$, and consequently, a low  $\mathcal{A}^{(U)}_\text{adv}$ of 12.6\% against the union of the perturbations. In contrast, FreeAdv+SNAP improves $\mathcal{A}^{(\ell_2)}_\text{adv}$ and $\mathcal{A}^{(\ell_1)}_\text{adv}$ by $17\%$ and $22\%$, respectively, accompanied by a 5\% improvement in $\mathcal{A}_\text{nat}$ and a small $2\%$ loss in $\mathcal{A}^{(\ell_\infty)}_\text{adv}$. This results in an overall robustness improvement of $20\%$  against the union of the perturbation models, setting a first benchmark for ResNet-50 on ImageNet. Upon increasing the network to ResNet-101, both natural and adversarial accuracies improve by $\approx4\%$ for FreeAdv, a trend also observed by \citet{shafahi2019adversarial}. SNAP further improves FreeAdv's results for $\mathcal{A}_\text{nat}$ and  $\mathcal{A}^{(U)}_\text{adv}$ by 4.3\% and 19.3\%. %

%% file: discussion.tex
\vspace{-0.2cm}
\section{Discussion}
\label{sec::Disc}
Given the wide popularity of $\ell_\infty$-AT, in this paper, we propose SNAP as an augmentation that generalizes the effectiveness of $\ell_\infty$-AT to the union of $(\ell_\infty, \ell_2, \ell_1)$ perturbations. SNAP's strength is  its simplicity and efficiency. Consequently, this work sets a first benchmark for ResNet-50 and ResNet-101 networks which are resilient to the union of $(\ell_\infty, \ell_2, \ell_1)$ perturbations on ImageNet. Note that norm-bounded perturbations  include a large class of attacks, \eg,  gradient-based~\cite{madry2018towards,rony2019decoupling,tramer2019adversarial,maini2019adversarial,chen2018ead,moosavi2016deepfool}, decision-based~\citep{brendel2018decision} and black-box~\cite{andriushchenko2020square} attacks. %

More work is needed to extend the proposed SNAP technique to %
attacks beyond norm-bounded additive perturbations, \eg, functional~\cite{laidlaw2019functional,xiao2018spatially}, rotation~\cite{engstrom2019exploring}, texture~\cite{bhattad2019unrestricted}, etc. We  provide preliminary evaluations in this direction in the Appendix. It is important to note that SNAP is meant to be an efficient technique for improving $\ell_\infty$-AT, and \emph{not} a new defense. Indeed defending against a large variety of attacks simultaneously remains  an open problem, with encouraging results from recent efforts~\cite{maini2019adversarial,laidlaw2021perceptual}.

Another limitation of our approach is that its benefits are demonstrated empirically. It is an inevitable consequence of a lack of any theoretical guarantees for underlying AT frameworks. An interesting direction of future work is to explore whether any theoretical guarantees can be derived for anisotropic shaped noise distributions in SNAP by building upon the recent developments in randomized smoothing~\cite{salman2019provably,yang2020randomized}. This could be a potential avenue for bridging the gap between certification bounds and empirical adversarial accuracy.

Finally, we believe that any effort on improving adversarial robustness of deep nets has net positive societal impact. However, recent past in this field has shown that any improvements in defense techniques also lead to more effective threat models. While such a cat-and-mouse game is of great intellectual value in the academic setting, it does have an unintentional negative societal consequence of equipping malicious outside actors  with a broad set of tools. This further underscores the well-recognized need for provable defenses.

%% file: Supplementary.tex
\begin{appendices}

\section{Robustness Stress Tests}
\label{app::StressTest}
We conduct robustness stress tests to confirm that the benefits of SNAP are sustained for a range of attack  norm-bounds, larger number of attack steps, and even for ``gradient-free'' attacks. For these experiments, we consider networks trained using TRADES and TRADES+SNAP (rows in Table~2 of the main paper), since they achieve the highest $\mathcal{A}^{(U)}_{\text{adv}}$ among the four SOTA AT frameworks.

\subsection{Sweeping norm-bounds and number of attack steps}

We sweep the number of PGD attack steps ($K$) and norm-bounds ($\epsilon$) for all three perturbations $(\ell_\infty,\ell_2,\ell_1)$ to confirm that the robustness gains from SNAP are achieved for a wider range of  attack norm bounds, and are sustained even after increasing attack steps. 
\begin{figure}[t]
	\begin{center}
		\includegraphics[width=0.75\linewidth]{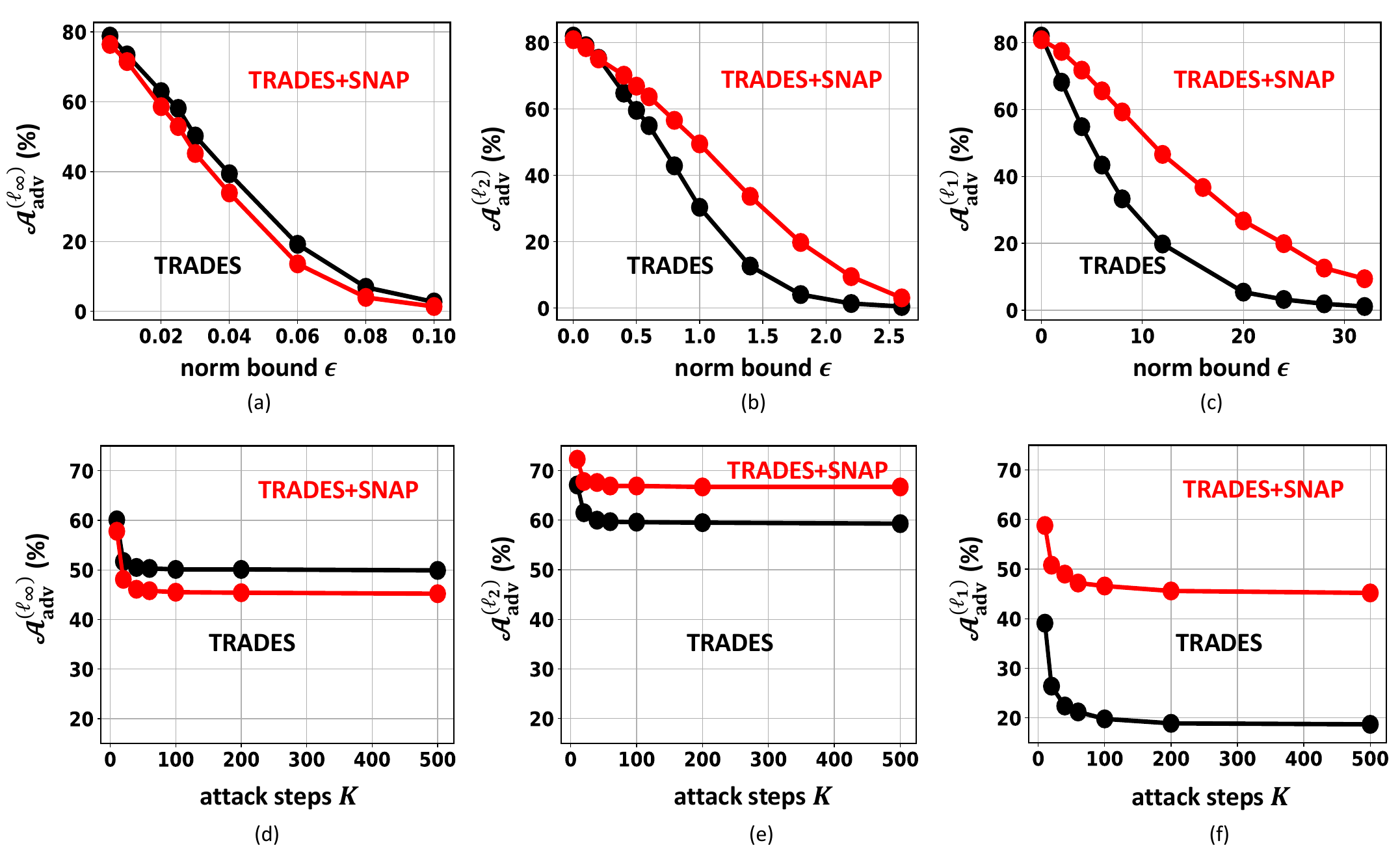}
	\end{center}
	\vspace{-0.4cm}
	\caption{ResNet-18 CIFAR-10 results: Adversarial accuracy \vs norm bound $\epsilon$ for: (a) $\ell_\infty$, (b) $\ell_2$, (c) $\ell_1$ PGD-100 attack. Adversarial accuracy \vs attack steps $K$ for (d) $\ell_\infty$ ($\epsilon=0.031$), (e) $\ell_2$ ($\epsilon=0.5$), (f) $\ell_1$ ($\epsilon=12$) PGD-100 attacks.
	}
	\label{fig::SuppStepsSwp} 
	\vspace{-0.25cm}
\end{figure}


Fig.~\ref{fig::SuppStepsSwp}(a)-(c) validates the main text Table~2 conclusion that TRADES+SNAP achieves large gains ($\sim20\%$) in $\mathcal{A}^{(\ell_1)}_{\text{adv}}$ and $\mathcal{A}^{(\ell_2)}_{\text{adv}}$ with a small ($\sim4\%$) drop in $\mathcal{A}^{(\ell_\infty)}_{\text{adv}}$. Furthermore, this conclusion holds for a large range of $\epsilon$ values for all three perturbations. Additionally, the gain in $\mathcal{A}^{(\ell_2)}_{\text{adv}}$ due to SNAP at $\epsilon=1.2$ is greater than the one reported in Table~2 for $\epsilon=0.5$.

Now we increase the attack steps $K$ to 500 and observe the impact on adversarial accuracy against $(\ell_\infty,\ell_2,\ell_1)$ perturbations in Fig.~\ref{fig::SuppStepsSwp}(d,e,f), respectively. In all cases, we observe hardly any change of the adversarial accuracy beyond $K=100$. Hence, as noted in the main text, we have chosen $K=100$ for all our experiments in the main text and in this supplementary. 

Recall we employ 10 random restarts as recommended by \citet{maini2019adversarial}  for \emph{all} our adversarial accuracy evaluations on CIFAR-10 data.

\subsection{Evaluating robustness against new attacks}
We evaluate adversarial accuracy against the recent DDN~\cite{rony2019decoupling}, Boundary~\cite{brendel2018decision}, and Square~\cite{andriushchenko2020square} attacks. The DDN attack was shown to be one of the SOTA gradient-based attacks, while boundary attack is one of the strongest ``gradient-free'' attacks. Of all the attacks considered in \citet{maini2019adversarial}, PGD turns out to be the strongest for $\ell_\infty$ and $\ell_1$ perturbations. Hence, in this section, we evaluate against $\ell_2$ norm-bounded DDN, boundary, and Square attacks. 

Following \citet{maini2019adversarial}, we use the FoolBox~\cite{rauber2017foolboxnative} implementation of the boundary attack, which uses 25 trials per iteration. For the DDN attack, we use 100 attack steps with appropriate logit averaging for $N_0=8$ noise samples \emph{before} computing the gradient in each step (similar to our PGD attack implementations). As mentioned in the main text, it eliminates any gradient obfuscation due to the presence of noise. 
\begin{table}[t]
	\centering
	\resizebox{0.7\columnwidth}{!}{
	\begin{tabular}{|c|>{\centering\arraybackslash}p{3cm}|>{\centering\arraybackslash}p{3cm}| } %
		\hline
		 &  {\makecell{TRADES} } & {\textbf{TRADES+SNAP }} \\ 
		\hline
		{\makecell{Natural Accuracy}}  & \textbf{82.1} & {80.9}    \\
    	\hline
		 \makecell{DDN~{\small\cite{rony2019decoupling}} {\small ($\epsilon=0.5$)} }  & {59.7}  & \textbf{65.8}  \\
		 \hline
		 \makecell{Boundary~{\small\cite{brendel2018decision}} {\small ($\epsilon=0.5$)} }  & {63.5}  & \textbf{67.0}  \\
		 \hline
		 \makecell{Square~{\small\cite{andriushchenko2020square}} {\small ($\epsilon=0.5$)} }  & {68.2}  & \textbf{72.7}  \\
		 \hline
	\end{tabular}}
	\vspace{0.2cm}
	\caption{ResNet-18 CIFAR-10 results showing natural accuracy (\%) and adversarial accuracy (\%) against $\ell_2$ norm bounded DDN attack~\cite{rony2019decoupling}, boundary attack~\cite{brendel2018decision}, and Square~\cite{andriushchenko2020square} for TRADES and TRADES+SNAP networks from Table~2 in the main text.} %
	\label{tab::MoreAttck}
	\vspace{-0.3cm}
\end{table}

Table~\ref{tab::MoreAttck} shows that SNAP improves adversarial accuracy against the DDN attack by $\sim 6\%$. This is similar to improvements seen against $\ell_2$-PGD attack in Table~2 in the main text. Similarly, TRADES+SNAP achieves $3.5\%$ ($4.5\%$) higher adversarial accuracy than TRADES against the Boundary~\cite{brendel2018decision} (Square~\cite{andriushchenko2020square}) attack.

\section{Additional Results}

\subsection{Comparison with Randomized Smoothing (RS)}
\label{app::RandSmooth}
\begin{figure}[t]
	\begin{center}
		\includegraphics[width=0.75\linewidth]{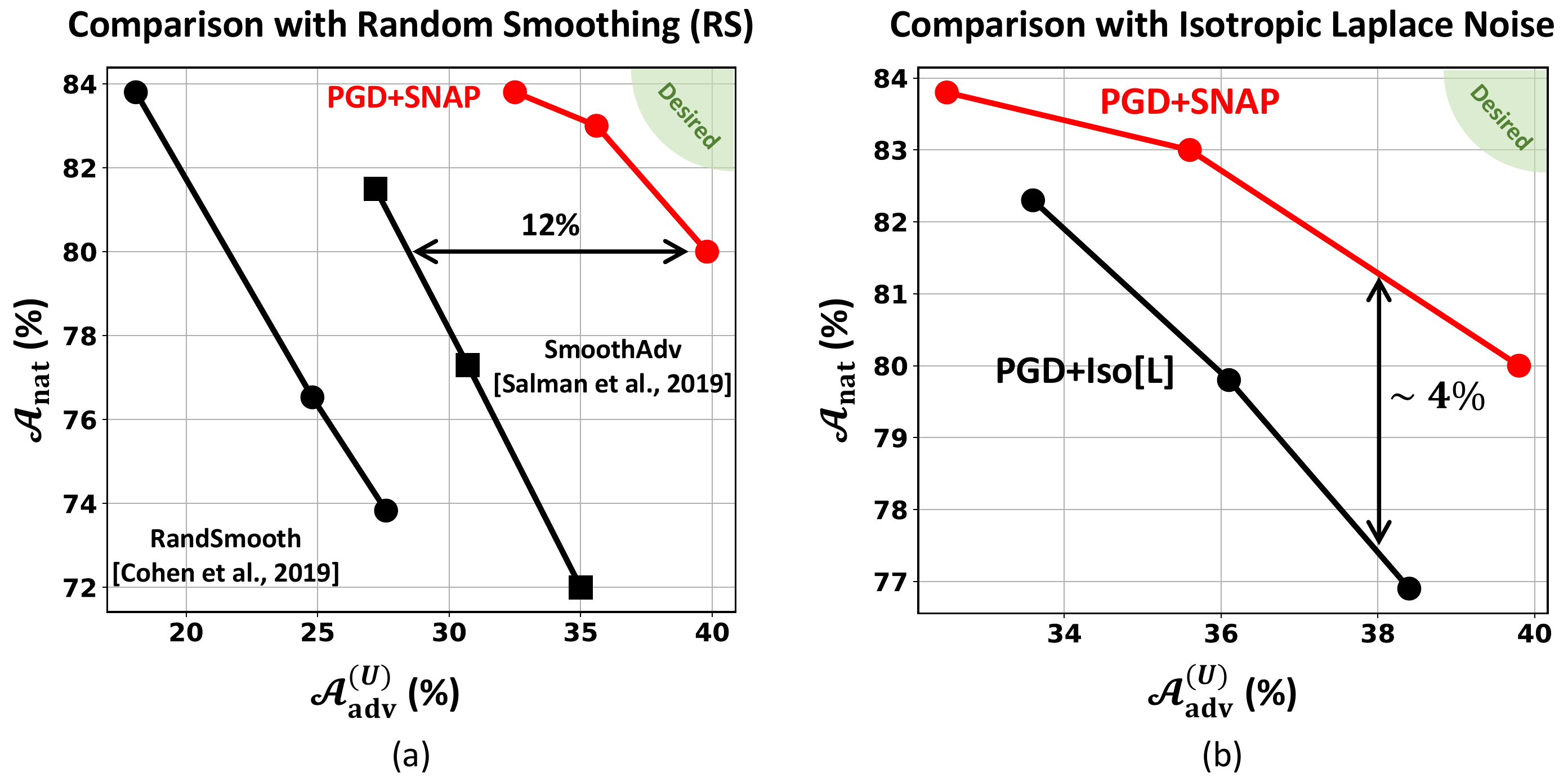}
	\end{center}
	\vspace{-0.4cm}
	\caption{ResNet-18 CIFAR-10 results: (a) $\mathcal{A}_\text{nat}$ \vs $\mathcal{A}^{(U)}_\text{adv}$ for RandSmooth~\cite{cohen2019certified},  SmoothAdv~\cite{salman2019provably}, and PGD+SNAP; (b) $\mathcal{A}_\text{nat}$ \vs $\mathcal{A}^{(U)}_\text{adv}$ for PGD+SNAP and PGD+Iso[L], where Iso[L] denotes a baseline SNAP alternative employing isotropic Laplace noise augmentation, \ie, without noise shaping. PGD+SNAP achieves better $\mathcal{A}_\text{nat}$ \vs $\mathcal{A}^{(U)}_\text{adv}$ trade-off due to noise shaping.
	}
	\label{fig::RSComp} 
	\vspace{-0.25cm}
\end{figure}

In this subsection, we compare with two SOTA randomized smoothing (RS) works, namely, RandSmooth~\cite{cohen2019certified}, and  SmoothAdv~\cite{salman2019provably}. They employ isotropic Gaussian noise. 
In Fig.~\ref{fig::RSComp}(a), we find that PGD+SNAP achieves a better $\mathcal{A}_\text{nat}$ \vs $\mathcal{A}^{(U)}_\text{adv}$ trade-off compared to both RandSmooth~\cite{cohen2019certified}, and  SmoothAdv~\cite{salman2019provably}. Specifically, note that SmoothAdv~\cite{salman2019provably} can also be viewed as isotropic Gaussian augmentation of $\ell_2$-PGD AT. Importantly, PGD+SNAP achieves a 12\% higher $\mathcal{A}^{(U)}_\text{adv}$ for the same $\mathcal{A}_\text{nat}$. This demonstrates the efficacy of \emph{shaped noise} in SNAP, which enhances the robustness to the union of $(\ell_\infty,\ell_2,\ell_1)$ perturbations. 

In order to further quantify importance of \emph{noise shaping}, we also compare $\ell_\infty$-PGD+SNAP with $\ell_\infty$-PGD+Iso[L], a stronger baseline alternative consisting of \emph{isotropic} Laplace noise augmentation, \ie, \emph{without any noise shaping}. Specifically, in Iso[L], the noise standard deviation is \emph{identical} in each direction, \ie, $\Sigma=\text{Diag}\bigg[\sqrt{\frac{P_\text{noise}}{D}},\dots,\sqrt{\frac{P_\text{noise}}{D}}\bigg]$. Note that such distributions have recently been explored for RS~\cite{yang2020randomized}. 

{Fig.~\ref{fig::RSComp}(b) plots the  $\mathcal{A}_\text{nat}$ \vs  $\mathcal{A}^{(U)}_\text{adv}$ trade-off for PGD+SNAP (\emph{red curve}) and PGD+Iso[L] (\emph{black curve}) by sweeping  $P_\text{noise}$. We find that PGD+SNAP achieves a better $\mathcal{A}_\text{nat}$ \vs  $\mathcal{A}^{(U)}_\text{adv}$ trade-off compared to PGD+Iso[L] by making more efficient use of noise power via noise shaping. Specifically, for $\mathcal{A}^{(U)}_\text{adv}\approx 38$, PGD+SNAP achieves a $\sim 4\%$ higher $\mathcal{A}_\text{nat}$. } 

\subsection{Comparison with \citet{madaan2020learning}}

\begin{table}[t]
	\centering
	\resizebox{0.85\columnwidth}{!}{
	\begin{tabular}{|c|>{\centering\arraybackslash}p{1.2cm}|>{\centering\arraybackslash}p{1.2cm}>{\centering\arraybackslash}p{1.2cm}>{\centering\arraybackslash}p{1.2cm}|>{\centering\arraybackslash}p{2.4cm}| } %
		\hline
		{\makecell{Method}} &  {\makecell{$\mathcal{A}_\text{nat}$} } & {\makecell{$\mathcal{A}^{(\ell_\infty)}_\text{adv}$ \\ {\scriptsize $\epsilon=0.03$} }} & {\makecell{$\mathcal{A}^{(\ell_2)}_\text{adv}$\\ {\scriptsize $\epsilon=0.31$} }} &  {\makecell{$\mathcal{A}^{(\ell_1)}_\text{adv}$\\ {\scriptsize $\epsilon=8$}  }} & {\makecell{Time per Epoch \\ (seconds)} } \\ [1.2ex]
		\hline
		{{MNG~\cite{madaan2020learning}}}  & {79.8} & {43.9} & \textbf{75.8} & {53.8} & $354^{\dagger}$  \\
		\hline
		{\textbf{PGD+SNAP}}  & \textbf{83.1} & \textbf{45.9} & 74.1 & \textbf{58.3} & \textbf{240}    \\
		\hline
	\end{tabular}}
	\vspace{0.1cm}
	\caption{ResNet-18 CIFAR-10 results showing a comparison between MNG~\cite{madaan2020learning} and PGD+SNAP (from Table~2 in the main text). All MNG numbers are exactly as reported in their paper. We reevaluate PGD+SNAP with our PGD attacks using the new $\epsilon$ values used by \citet{madaan2020learning}. PGD+SNAP achieves 3\%, 2\%, 4.5\% higher $\mathcal{A}_\text{nat}$, $\mathcal{A}^{(\ell_\infty)}_\text{adv}$, $\mathcal{A}^{(\ell_1)}_\text{adv}$, respectively, while being at least $\sim40\%$ faster in terms of epoch time. $\dagger$: Note that MNG time is measured on NVIDIA GeForce RTX 2080Ti (by \citet{madaan2020learning}), while PGD+SNAP is measured on NVIDIA Tesla P100. An RTX 2080Ti has \emph{20\% more} CUDA cores than a Tesla P100. } %
	\label{tab::MadaanComp}
\end{table}

The meta-noise generator (MNG)~\cite{madaan2020learning} employs a multi-layer deep-net to generate noise samples during AT. Importantly, MNG still employs multiple attacks during  training, but samples only one of the attacks randomly at a time to reduce the training cost. 

However, they have yet to release their code or pretrained models even though their work was posted on arXiv a year ago.
Absence of public codes from \citet{madaan2020learning} makes it  difficult to clearly compare with their work, especially in terms of training time. Nonetheless, in this subsection, we try our best to ensure that the comparison is fair. Table~\ref{tab::MadaanComp} reports natural and adversarial accuracy of MNG against $(\ell_\infty, \ell_2, \ell_1)$ attacks as reported by \citet{madaan2020learning}. We find that PGD+SNAP achieves 3\%, 2\%, 4.5\% higher $\mathcal{A}_\text{nat}$, $\mathcal{A}^{(\ell_\infty)}_\text{adv}$, and  $\mathcal{A}^{(\ell_1)}_\text{adv}$, respectively. Note that \citet{madaan2020learning} evaluate $\mathcal{A}^{(\ell_\infty)}_\text{adv}$ and $\mathcal{A}^{(\ell_2)}_\text{adv}$ against PGD-50 attacks, whereas here we employ PGD-100 attacks and, following their protocol,  evaluate on the entire CIFAR-10 dataset with a single restart. Furthermore, epoch time for PGD+SNAP is $1.4\times$ smaller than that of MNG~\cite{madaan2020learning} even though MNG time was measured on a more recent NVIDIA RTX 2080Ti, which has 20\% more CUDA cores than the Tesla P100 GPU that we used for PGD+SNAP. 

Importantly, a key advantage of SNAP is its scalability. We are able to report robust ResNet-50 and ResNet-101 networks on ImageNet (Table~4 in the main text), whereas \citet{madaan2020learning} report results only up to $64\times 64$ TinyImageNet.

\subsection{SVHN results}

\begin{table}[t]
	\centering
	\resizebox{0.7\columnwidth}{!}{
	\begin{tabular}{|c|>{\centering\arraybackslash}p{0.9cm}|>{\centering\arraybackslash}p{1cm}>{\centering\arraybackslash}p{0.9cm}>{\centering\arraybackslash}p{0.9cm}|>{\centering\arraybackslash}p{0.9cm}| } %
		\hline
		{\makecell{Method}} &  {\makecell{$\mathcal{A}_\text{nat}$} } & {\makecell{$\mathcal{A}^{(\ell_\infty)}_\text{adv}$ \\ {\scriptsize $\epsilon=0.03$} }} & {\makecell{$\mathcal{A}^{(\ell_2)}_\text{adv}$\\ {\scriptsize $\epsilon=0.5$} }} &  {\makecell{$\mathcal{A}^{(\ell_1)}_\text{adv}$\\ {\scriptsize $\epsilon=8$}  }} & {\makecell{$\mathcal{A}^{(U)}_\text{adv}$} } \\ [1.2ex]
		\hline
		{\makecell{PGD}}  & \textbf{89.9} & \textbf{45.3} & 34.9 & {4.8} & {4.8}   \\
		\hline
		{\textbf{PGD+SNAP}}  & {89.3} & {44.0} & \textbf{67.4} & \textbf{48.3} & \textbf{36.3}   \\
		\hline
	\end{tabular}}
	\vspace{0.2cm}
	\caption{ResNet-18 SVHN results showing the impact of SNAP augmentation of $\ell_\infty$-PGD~\cite{madry2018towards} AT frameworks. Adding SNAP improves $\mathcal{A}^{(U)}_\text{adv}$ by $\sim 30\%$ while having only a small impact on $\mathcal{A}_\text{nat}$ and $\mathcal{A}^{(\ell_\infty)}_\text{adv}$.} %
	\label{tab::SVHNResults}
	\vspace{-0.3cm}
\end{table}

Table~\ref{tab::SVHNResults} shows PGD and PGD+SNAP results on SVHN data. We train both PGD and PGD+SNAP models for  100 epochs using a piece-wise LR schedule. We start with an initial LR of 0.01 and decay it once at the 95th epoch.  

In Table~\ref{tab::SVHNResults}, we observe a trend that is similar to our observations for CIFAR-10 and ImageNet results. In particular, for SVHN, SNAP turns out to be even more effective, with $\sim 30\%$ improvement in $\mathcal{A}^{(U)}_\text{adv}$ while almost preserving both $\mathcal{A}_\text{nat}$ and $\mathcal{A}^{(\ell_\infty)}_\text{adv}$. 

\subsection{Impact of SNAP on prediction complexity}

\begin{table}[t]
	\centering
	\resizebox{0.3\columnwidth}{!}{
	\begin{tabular}{|c|>{\centering\arraybackslash}p{1.9cm}| } 
		\hline
		{\makecell{Method}} &  {\makecell{$\mathcal{A}_\text{nat}$ (\%)} } \\ [1.2ex]
		\hline
		{\makecell{TRADES}}  & {81.7}   \\
		\hline
		\multicolumn{2}{|c|}{\textbf{TRADES+SNAP}} \\
		\hline
		{$N_0=1$}  & {80.1{\small $\pm0.22$}}   \\
		\hline
		{$N_0=2$}  & {80.3{\small $\pm0.14$}}   \\
		\hline
		{$N_0=4$}  & {80.7{\small $\pm0.12$}}   \\
		\hline
		{$N_0=8$}  & {80.9{\small $\pm0.10$}}   \\
		\hline
		{$N_0=16$}  & {80.9{\small $\pm0.08$}}   \\
		\hline
	\end{tabular}}
	\vspace{0.2cm}
	\caption{ResNet-18 CIFAR-10 results showing SNAP's impact on the prediction complexity, where $N_0$ denotes the number of noise samples employed to estimate $\mathbb{E}[\cdot]$ in Eq.~$(2)$ in the main text. We find that for mere accuracy estimation, even a single forward pass ($N_0=1$) suffices. ${\small \pm \text{xx}}$ denotes the standard deviation over 10 independent test runs.} %
	\label{tab::PredTimeComp}
	\vspace{-0.3cm}
\end{table}

While SNAP augmentation has a modest impact on the training time (Table~3 in the main text), here we check whether it could \emph{potentially} increase the model prediction complexity due to the need to estimate the expectation  $\mathbb{E}[\cdot]$ in Eq.~$(2)$ in the main text. 

As expected, by increasing $N_0$, the deviation of the $\mathcal{A}_\text{nat}$ estimate reduces (see Table~\ref{tab::PredTimeComp}). However, we find that for accuracy estimation, a single forward pass ($N_0=1$) suffices. Specifically, an  $\mathcal{A}_\text{nat}$ estimate with $N_0=1$ is within 1\% of the $\mathcal{A}_\text{nat}$ estimate with $N_0=16$. Furthermore, even with $N_0=1$, the standard deviation of $\mathcal{A}_\text{nat}$ is as low as $\sim0.2$\%. Thus, the impact of SNAP on prediction complexity can be very small. 

\subsection{Subspace analysis of adversarial perturbations for  TRADES+SNAP model}

\begin{figure}[t]
	\begin{center}
		\includegraphics[width=0.7\linewidth]{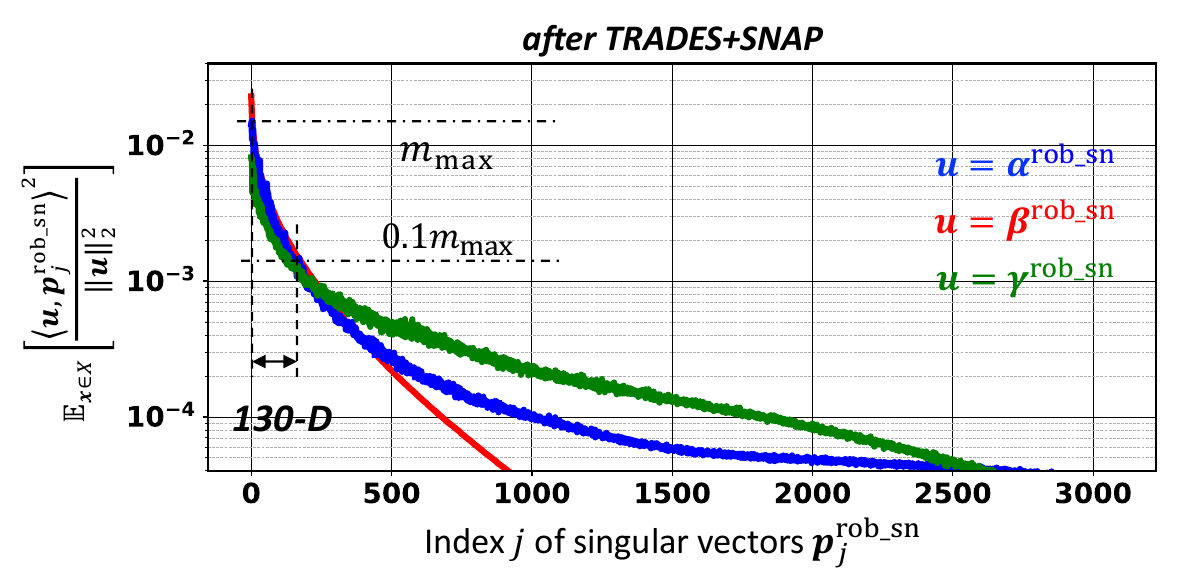}
	\end{center}
	\vspace{-0.1cm}
	\caption{Normalized mean squared projections of three perturbation types on the singular vector basis $\mathcal{P}^{\kappa}$ of $\ell_2$ perturbations of ResNet18 on CIFAR-10 after TRADES+SNAP training ($\kappa\equiv\text{rob\_sn}$). The singular vectors $\vp_i^{\kappa}$ comprising $\mathcal{P}^{\kappa}=\{\vp_1^{\kappa},\dots,\vp_D^{\kappa}\}$ are ordered in descending order of their singular values.}
	\label{fig::SnapSubspace} 
	\vspace{-0.25cm}
\end{figure}

In this subsection, we carry out a subspace analysis of adversarial perturbations (Section~3 in the main text) for TRADES+SNAP. We confirm that our hypothesis in Section~3 holds even after SNAP augmentation of TRADES. Following the same experimental setup and the notation from Section~3 in the main text, we compute perturbations $\bm{\alpha}_i$, $\bm{\beta}_i$, and $\bm{\gamma}_i$ for each $\vx_i\in X$ for  ResNet-18 trained using TRADES+SNAP, \ie, $\kappa\equiv\text{rob\_sn}$. We compute the singular vector basis  $\mathcal{P}^{\kappa}$ for the set of $\ell_2$ bounded perturbations $\Delta^{\kappa} =\{\bm{\beta}_1^{\kappa},\dots,\bm{\beta}_{|X|}^{\kappa}\}$. Fig.~\ref{fig::SnapSubspace} plots the normalized mean squared projections of the three types of perturbation vectors on the singular vector basis $\mathcal{P}^{\kappa}$ of a TRADES+SNAP trained ResNet-18. We find that the projections generally follow the same trend as those for a TRADES-trained network which are shown in Fig.~3(b) of the main text. However, we also notice that after SNAP augmentation, the three perturbation types get squeezed into an even smaller 130-dimensional subspace, \ie, projections are $<10\%$ of the maximum projection value for all dimensions beyond the first 130 dimensions.

\begin{table}[t]
	\centering
	\resizebox{0.7\columnwidth}{!}{
	\begin{tabular}{|c|>{\centering\arraybackslash}p{0.9cm}|>{\centering\arraybackslash}p{1cm}>{\centering\arraybackslash}p{0.9cm}>{\centering\arraybackslash}p{0.9cm}|>{\centering\arraybackslash}p{0.9cm}| } %
		\hline
		{\makecell{Method}} &  {\makecell{$\mathcal{A}_\text{nat}$} } & {\makecell{$\mathcal{A}^{(\ell_\infty)}_\text{adv}$ \\ {\scriptsize $\epsilon=0.03$} }} & {\makecell{$\mathcal{A}^{(\ell_2)}_\text{adv}$\\ {\scriptsize $\epsilon=0.5$} }} &  {\makecell{$\mathcal{A}^{(\ell_1)}_\text{adv}$\\ {\scriptsize $\epsilon=12$}  }} & {\makecell{$\mathcal{A}^{(U)}_\text{adv}$} } \\ [1.2ex]
		\hline
		{\makecell{PGD}}  & {84.6} & {48.8} & 62.3 & {15.0} & {15.0}   \\
		\hline
		\hline
		\multicolumn{6}{|c|}{\textbf{Noise shaping basis $V=\mathbf{I}_{D\times D}$ }} \\
		\hline
		\hline
		 \textbf{+SNAP[G]}  & {80.7}  & \textbf{45.7}  & {66.9}  &  {34.6} &  {31.9} \\
		 \textbf{+SNAP[U]}  & \textbf{85.1}  & {42.7}  & \textbf{66.7}  &  {28.6} &  {26.6}  \\
		\textbf{+SNAP[L]}  & {83.0}  & {44.8}  & \textbf{68.6}  &  \textbf{40.1} &  \textbf{35.6} \\
    	\hline
		\hline
		\multicolumn{6}{|c|}{\textbf{Noise shaping basis $V=U_{\text{img}}$}} \\
		\hline
		\hline
		 \textbf{+SNAP[G]}  & {81.7}  & \textbf{48.9}  & {67.5}  &  \textbf{29.8} &  \textbf{28.7} \\
		 \textbf{+SNAP[U]}  & \textbf{82.0}  & {46.6}  & \textbf{67.8}  &  {27.8} &  {25.7}  \\
		\textbf{+SNAP[L]}  & {81.7}  & {46.8}  & {65.9}  &  {28.5} &  {27.4} \\

    	\hline %
	\end{tabular}}
	\vspace{0.2cm}
	\caption{ResNet-18 CIFAR-10 results showing the impact of noise shaping basis $V$ for $\ell_\infty$-PGD~\citep{madry2018towards} AT framework with SNAP. In this table, SNAP[G], SNAP[U], and SNAP[L] denote shaped noise augmentations with Gaussian, Uniform, and Laplace noise distributions, respectively, and $U_\text{img}$ refers to the singular vector basis of the training images.} %
	\label{tab::NoiseBasisComp}
	\vspace{-0.3cm}
\end{table}

\subsection{Impact of noise shaping in the image basis}
\label{subsec::NoiseBasis}
 Recall that, for all experiments in the main text, we chose the noise shaping basis $V=\mathbf{I}_{D\times D}$, \ie, the noise was shaped and added in the standard basis in $\mathbb{R}^D$, where $\mathbf{I}_{D\times D}$ denotes the identity matrix (see Eq.~(1) and Alg.~1 in the main text).

In this section, we explore the shaped noise augmentation in the \emph{image basis}, \ie, singular vector basis of the training set images. Specifically, we choose $V=U_\text{img}=[\vu_1,\dots,\vu_D]$, where $U_\text{img}$ denotes the singular vector basis of the images in the training set. Thus, the sampled noise vector $\rvn_0$ (see Eq.~(1) in the main text) is scaled by direction-wise standard deviation matrix $\Sigma$ and \emph{rotated} by $U_\text{img}$ before being added to the input image $\vx$. 

The rationale for choosing $V=U_\text{img}$ is as follows: Recent works~\cite{ilyas2019adversarial,shafahi2019adversarial,santurkar2019image} have demonstrated the generative behavior of adversarial perturbations of  networks trained with single-attack AT, \ie, adversarial perturbations of robust networks exhibit semantics similar to the input images. Thus, the perturbation basis (see section~3 in the main text) of the robust networks trained with single-attack AT seems to be aligned with the image basis. 

We repeat the experiments in Table~1 of the main text while keeping all the settings  \emph{identical} except for choosing $V=U_\text{img}$ instead of $V=\mathbf{I}_{D\times D}$. Table~\ref{tab::NoiseBasisComp} shows the results. The first three rows correspond to $V=\mathbf{I}_{D\times D}$ and are reproduced from Table~1 of the main text. Note that, in order to preserve $\mathcal{A}_\text{nat}> 81\%$, we need to reduce $P_\text{noise}=60$ when $V=U_\text{img}$, since the noise is now pixel-wise correlated.

In Table~\ref{tab::NoiseBasisComp}, we notice that $\mathcal{A}^{(\ell_1)}_{\text{adv}}$ is significantly reduced when $V=U_\text{img}$ as compared to the case $V=\mathbf{I}_{D\times D}$. 
More interestingly, all three types of noise distributions result in similar values for $\mathcal{A}^{(\ell_1)}_{\text{adv}}$ when $V=U_\text{img}$.
We discuss this phenomenon in the next section, \ie, Sec.~\ref{app::NoiseHist} below. 

Table~\ref{tab::NoiseBasisComp} shows that the orientation of a noise vector is as important as its distribution. The simpler choice of $V=\mathbf{I}_{D\times D}$ turns out to be more effective.

\subsection{Understanding the effectiveness of SNAP[L] for $\ell_\infty$ AT }
\label{app::NoiseHist}
In this subsection, we conduct additional studies to further understand the following two observations in SNAP: (i) shaped Laplace noise is particularly effective (Table~1 in the main text), and (ii) rotating noise vectors ($V=U_\text{img}$) reduces their effectiveness (Table~\ref{tab::NoiseBasisComp} in this Supplementary). We study the properties of the noise vector $\rvn$ for different noise distributions.

We conjecture that the Laplace distribution is most effective because of its heavier tail compared to Gaussian and Uniform distributions of the same variance. A long-tailed distribution will generate more large magnitude elements in a vector drawn from it and hence is more effective in emulating a strong $\ell_1$-norm bounded perturbation. Furthermore, the standard (un-rotated) basis preserves this unique attribute of samples drawn from such distributions.

\begin{figure}[t]
	\begin{center}
		\includegraphics[width=0.75\linewidth]{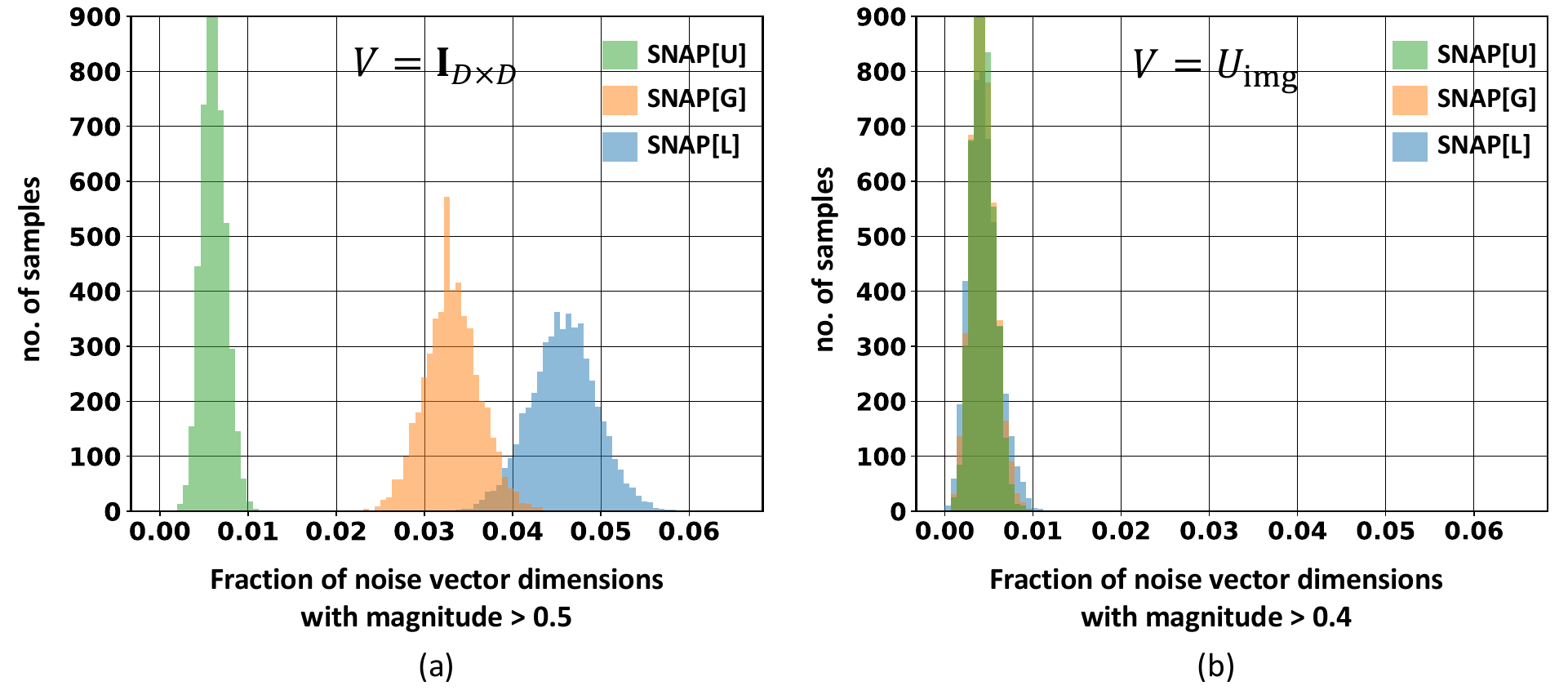}
	\end{center}
	\vspace{-0.4cm}
	\caption{ResNet18 CIFAR-10 results: histograms of the fraction of noise vector dimensions with magnitude (a) $>0.5$ when $V=\mathbf{I}_{D\times D}$, and (b) $>0.4$ when $V=U_{\text{img}}$. Histograms are plotted for 5000 random noise samples $\rvn$. The three shaped noise distributions are from the corresponding networks in Table~\ref{tab::NoiseBasisComp}.}
	\label{fig::NoiseStdHist} 
	\vspace{-0.25cm}
\end{figure}

This conjecture is validated by Fig.~\ref{fig::NoiseStdHist}(a) which shows that noise samples drawn from the Laplace distribution in the standard basis have the highest average number of dimensions with large ($>0.5$) magnitudes, followed by Gaussian and Uniform distributions. This correlates well with the results in Table~1 in the main text and Table III (first three rows), in that $\mathcal{A}^{(\ell_1)}_\text{adv}$ is the highest for Laplace followed by those for Gaussian and Uniform. Additionally, the use of $V=U_\text{img}$ dissolves this distinction between the three distributions as shown in Fig.~\ref{fig::NoiseStdHist}(b) which explains the similar (and lower)  $\mathcal{A}^{(\ell_1)}_\text{adv}$ values for all three distributions in Table~III.

Thus, we confirm that the type of noise plays an important role in robustifying single-attack $\ell_\infty$ AT frameworks to the union of multiple perturbation models. Specifically, the noise vectors with higher fraction of noise dimensions with larger magnitudes are better at complementing $\ell_\infty$ AT frameworks.

\subsection{Evaluating common corruptions and functional attack}

\begin{table}[t]
	\centering
	\resizebox{0.6\columnwidth}{!}{
	\begin{tabular}{|c|>{\centering\arraybackslash}p{1.2cm}|>{\centering\arraybackslash}p{1.2cm}|>{\centering\arraybackslash}p{1.5cm}|>{\centering\arraybackslash}p{1.9cm}| } %
		\hline
		\multirow{2}{4em}{\makecell{Method}} &  \multirow{2}{4em}{\makecell{$\mathcal{A}_\text{nat}$} } & \multirow{2}{4em}{\makecell{$\mathcal{A}^{(U)}_\text{adv}$} } & \multirow{2}{4em}{\makecell{ $\mathcal{A}_\text{cc}$}} & {\makecell{$\mathcal{A}^{\text{(f)}}_\text{adv}$} } \\
		
		& & & & \makecell{ReColorAdv} \\ [1.2ex]
		\hline
		{\makecell{Vanilla}}  & \textbf{94.5} & {0.0} & 72.0 & {0.9}   \\
		\hline
		\hline
		{\makecell{$\ell_\infty$-PGD}}  & {84.6} & {15.0} & \textbf{75.6} & {53.5}   \\
		\hline
		\hline
		\multicolumn{5}{|c|}{\textbf{Noise shaping basis $V=\mathbf{I}_{D\times D}$ }} \\
		\hline
		\hline
		 \textbf{+SNAP[G]}  & {80.7}&  {31.9}  & 72.8 &  \textbf{55.1} \\
		 \textbf{+SNAP[U]}  & \textbf{85.1}&  {26.6}  & 75.0 &  {46.9}  \\
		\textbf{+SNAP[L]}  & {83.0}&  \textbf{35.6}  & 75.3 & 51.3 \\
    	\hline
		\hline
		\multicolumn{5}{|c|}{\textbf{Noise shaping basis $V=U_{\text{img}}$}} \\
		\hline
		\hline
		 \textbf{+SNAP[G]}  & {81.7}&  {28.7}  & 73.6 &  {54.5} \\
		 \textbf{+SNAP[U]}  & {82.0}&  {25.7}  & 73.1 &  {54.0}  \\
		\textbf{+SNAP[L]}  & {81.7}&  {27.4}  & 73.4 &  \textbf{55.3} \\
    	\hline %
	\end{tabular}}
	\vspace{0.2cm}
	\caption{ResNet-18 CIFAR-10 results showing natural accuracy $\mathcal{A}_\text{nat}$, adversarial accuracy $\mathcal{A}^{(U)}_\text{adv}$ against the union of $(\ell_\infty,\ell_2,\ell_1)$ perturbations, accuracy $\mathcal{A}_\text{cc}$ in the presence of common corruptions~\cite{hendrycks2018benchmarking}, and adversarial accuracy $\mathcal{A}^{\text{(f)}}_\text{adv}$ against a functional adversarial attack ReColorAdv~\cite{laidlaw2019functional}. All accuracy numbers are in \%. In this table, $\mathbf{I}_{D\times D}$ denotes $D$-dimensional identity matrix, while  $U_\text{img}$ denotes singular vector basis of the training images. We find that SNAP augmentations of $\ell_\infty$-PGD significantly ($\approx 20\%$) improve $\mathcal{A}^{(U)}_\text{adv}$ while preserving both $\mathcal{A}_\text{cc}$ and $\mathcal{A}^{\text{(f)}}_\text{adv}$.} %
	\label{tab::CorrAccuracies}
	\vspace{-0.3cm}
\end{table}

In this subsection, we check if there are any other downsides of SNAP when it improves robustness against the union of $(\ell_\infty,\ell_2,\ell_1)$ perturbations. In particular, we check if SNAP improvements are achieved at the cost of a drop in accuracy against common corruptions~\cite{hendrycks2018benchmarking} or functional adversarial attacks~\cite{laidlaw2019functional}.

We use corrupted images provided by \citet{hendrycks2018benchmarking} to estimate accuracy in the presence of common corruptions  ($\mathcal{A}_\text{cc}$). We average the accuracy numbers across different corruption strengths and types. Also, we use the ReColorAdv setup  of \citet{laidlaw2021perceptual} to estimate accuracy  against functional adversarial attacks ($\mathcal{A}^{\text{(f)}}_\text{adv}$). We also  make it \emph{adaptive} to our defense framework via appropriate noise averaging (similar to our adaptive PGD attacks~\cite{salman2019provably} discussed in the main text) to eliminate any gradient obfuscations. 
As observed in Table~\ref{tab::CorrAccuracies}, SNAP augmentations of PGD AT generally preserve both $\mathcal{A}_\text{cc}$ and $\mathcal{A}^{\text{(f)}}_\text{adv}$. In particular, 20.6\% improvement in $\mathcal{A}^{(U)}_\text{adv}$ via PGD+SNAP[L] (with $V=\mathbf{I}_{D\times D}$) is accompanied with the same $\mathcal{A}_\text{cc}$ and only a $2.2\%$ lower $\mathcal{A}^{\text{(f)}}_\text{adv}$ ($=51.3\%$) compared to PGD AT. In contrast, vanilla training achieves an $\mathcal{A}^{\text{(f)}}_\text{adv}$ of only 0.9\%. Even  with $V=U_{\text{img}}$, PGD+SNAP[L] achieves a 1.8\% higher $\mathcal{A}^{\text{(f)}}_\text{adv}$ along with a 12.4\% improvement in $\mathcal{A}^{(U)}_\text{adv}$. Note that all $\mathcal{A}^{(U)}_\text{adv}$ numbers are idential to the ones reported in Sec.~\ref{subsec::NoiseBasis}.

We conclude that SNAP augmentation of PGD AT improves $\mathcal{A}^{(U)}_\text{adv}$ by up to 20\% while preserving its robustness against common corruptions and functional adversarial attacks. Thus, SNAP expands the capabilities of $\ell_\infty$ AT frameworks without any significant downside. However, further work is required to improve robustness to a larger class adversarial attacks, such as rotation~\cite{engstrom2019exploring}, texture~\cite{bhattad2019unrestricted}, etc., simultaneously. 

\subsection{Error bars}

In this subsection, we confirm that benefits of SNAP are not specific to any particular choice of random seed. Specifically, we run both PGD+SNAP (with superconvergence) and FreeAdv+SNAP (see Table~3 in the main text) training four times with different random seeds.
Table~\ref{tab::ErrorBars} shows the mean accuracy and its standard deviation for each of $\mathcal{A}_\text{nat}$, $\mathcal{A}^{(\ell_\infty)}_\text{adv}$, $\mathcal{A}^{(\ell_2)}_\text{adv}$,  $\mathcal{A}^{(\ell_1)}_\text{adv}$, and $\mathcal{A}^{(U)}_\text{adv}$ with ResNet-18 on CIFAR-10. We find that the standard deviation of accuracy is  $\approx0.5$\% in almost all cases. This demonstrates the ease of replicating SNAP results. 

\begin{table}[t]
	\centering
	\resizebox{\columnwidth}{!}{
	\begin{tabular}{|c|>{\centering\arraybackslash}p{1.6cm}|>{\centering\arraybackslash}p{1.6cm}>{\centering\arraybackslash}p{1.6cm}>{\centering\arraybackslash}p{1.6cm}|>{\centering\arraybackslash}p{1.6cm}| } %
		\hline
		{\makecell{Method}} &  {\makecell{$\mathcal{A}_\text{nat}$} } & {\makecell{$\mathcal{A}^{(\ell_\infty)}_\text{adv}$ \\ {\scriptsize $\epsilon=0.03$} }} & {\makecell{$\mathcal{A}^{(\ell_2)}_\text{adv}$\\ {\scriptsize $\epsilon=0.5$} }} &  {\makecell{$\mathcal{A}^{(\ell_1)}_\text{adv}$\\ {\scriptsize $\epsilon=12$}  }} & {\makecell{$\mathcal{A}^{(U)}_\text{adv}$} } \\ [1.2ex]
		\hline
		{\textbf{PGD+SNAP}}  & {82.5$\pm$0.27} & {43.1$\pm$0.61} & 66.9$\pm$0.57 & {39.0$\pm$0.41} & {33.7$\pm$0.29}   \\
		\hline
		 {\textbf{FreeAdv+SNAP}}  & {83.4$\pm$0.25}  & {39.2$\pm$0.74}  & {65.7$\pm$0.55}  &  {36.5$\pm$0.60} &  {30.4$\pm$0.83} \\
		\hline
	\end{tabular}}
	\vspace{0.1cm}
	\caption{ResNet-18 CIFAR-10 results showing the mean and standard deviation for all accuracies over four different training runs of PGD+SNAP (with superconvergence) and FreeAdv+SNAP. As observed, the standard deviation in accuracy is $\approx0.5$\% in almost all cases, demonstrating the ease of replicating  SNAP results.} %
	\label{tab::ErrorBars}
\end{table}

\section{Additional Details}
\label{app::Details}

\subsection{Details of Hyperparameters}

\subsubsection{Attack hyperparameters}

As mentioned in the main text, we follow basic PGD attack formulations of \citet{maini2019adversarial}. We further enhance them to target the full defense -- SN layer -- since SNAPnet is end-to-end differentiable. Specifically, we backpropagate to the primary input $\vx$ through the SN layer (see Fig.~4(b) in the main text). Thus, the final shaped noise distribution is exposed to the adversary. We also account for the $\mathbb{E}_\rvn[\cdot]$ (see Eq.~(2) in the main text) by explicitly averaging deep net logits over $N_0$ noise samples \emph{before} computing the gradient, which eliminates any gradient obfuscation, and was shown to be the strongest attack against noise augmented models~\cite{salman2019provably}. We choose $N_0=8$ for all our attack evaluations. 

For $\ell_2$ and $\ell_\infty$ PGD attacks, we choose steps size $\alpha=0.1\epsilon$. For $\ell_1$ PGD attacks, we choose the exact same configuration as \citet{maini2019adversarial}. 

\subsubsection{Training hyperparameters}

As mentioned in the main text, we introduce SNAP without changing any hyperparameters of BASE() AT. All BASE() and BASE()+SNAP  training runs on CIFAR-10 employ an SGD optimizer with a fixed momentum of $0.9$, batch size of $250$, and weight decay of $2\times 10^{-4}$. Also, while accounting for the $\mathbb{E}_\rvn[\cdot]$ (see Eq.~(2) in the main text), note that $N_0=1$ suffices during BASE()+SNAP training. Below we provide specific details for each SOTA AT framework:

{\textbf{BASE() $\equiv$ PGD~\cite{madry2018towards} on CIFAR-10:}}

$\ell_\infty$-PGD AT employed $\ell_\infty$-bounded PGD-$K$ attack with  $\epsilon=0.031$, step size $\alpha=0.008$, and $K=10$. For $\ell_2$-PGD AT, we used an $\ell_2$-bounded PGD-$K$ attack with $\epsilon=0.5$, step size $\alpha=0.125$ and $K=10$. 
Following \citet{rice2020overfitting}, we employed $100$ epochs for PGD AT with step learning rate (LR) schedule, where LR was decayed from 0.1 to 0.01 at epoch 96. Following \citet{maini2019adversarial}, we also employed \emph{their} cyclic LR schedule to achieve superconvergence in 50 epochs. Following \citet{maini2019adversarial}, we set weight decay to $5\times 10^{-4}$ in PGD AT. 

In PGD+SNAP, the noise variances were updated every $U_f=10$ epochs and we use $P_\text{noise}=160$ in Tables~1,2, and 3 in the main text.

{\textbf{BASE() $\equiv$ TRADES~\cite{zhang2019theoretically} on CIFAR-10:}}

Following \citet{zhang2019theoretically}, TRADES AT employed $\ell_\infty$-bounded perturbations with $\epsilon=0.031$, step size $\alpha=0.007$, and attack steps $K=10$. We set TRADES parameter $1/\lambda=5$, which controls the weighing of its robustness regularizer. It was trained for 100 epochs with a step LR schedule, where LR was decayed to \{0.01,0.001,0.0001\} at the epochs $\{75,90,100\}$, respectively. Following \citet{maini2019adversarial}, we also employed \emph{their} cyclic LR schedule to achieve superconvergence in 50 epochs, while keeping all other settings identical. 

In TRADES+SNAP, the noise variances were updated every $U_f=10$ epochs and we use $P_\text{noise}=120$ in Tables~2 and 3 in the main text. 

{\textbf{BASE() $\equiv$ FreeAdv~\cite{shafahi2019adversarial} on CIFAR-10:}}

Following \citet{shafahi2019adversarial}, FreeAdv AT was trained for 25 epochs, each consisting of a replay of 8. It employed $\ell_\infty$ perturbations with $\epsilon=0.031$. The learning rate was decayed to $\{0.01,0.001,0.0001\}$ at  epochs $\{13,19,23\}$, respectively. 

In FreeAdv+SNAP, the noise variances were updated every $U_f=5$ epochs, since the replay of 8 scales down the total number of epochs. Also, we use $P_\text{noise}=160$ in Tables~2 and 3 in the main text.

{\textbf{BASE() $\equiv$ FastAdv~\cite{wong2020fast} on CIFAR-10:}}

Following \citet{wong2020fast}, FastAdv AT employed a single-step $\ell_\infty$ norm bounded FGSM  attack with $\epsilon=8/255$, step size $\alpha=10/255$, and random noise initialization. It was trained for 50 epochs with the \emph{same} cyclic LR schedule used by \citet{wong2020fast}. We used a weight decay of $5\times 10^{-4}$.

In FastAdv+SNAP, the noise variances were updated every $U_f=10$ epochs and we use $P_\text{noise}=200$ in Tables~2 and 3 in the main text. 
 
{\textbf{BASE() $\equiv$ FreeAdv~\cite{shafahi2019adversarial} on ImageNet:}}

Following \citet{shafahi2019adversarial}, FreeAdv AT was trained for 25 epochs, each consisting of a replay of 4. It employed $\ell_\infty$ perturbations with $\epsilon=4/255$, \emph{identical} to the authors' original setup. The LR was decayed by 0.1 every 8 epochs, starting with the initial LR of 0.1. We used weight decay of $1\times 10^{-4}$.

In FreeAdv+SNAP, the noise variances were updated every $U_f=5$ epochs, since the replay of 4 scales down the total number of epochs. Also, we use $P_\text{noise}=4500$ in Table~4 in the main text, which corresponds to noise standard deviation of $\sim 0.17$ per pixel \emph{on average}. 

{\textbf{MSD-$K$~\cite{maini2019adversarial} experiments for $K\in\{ 30, 20, 10, 5 \}$:}}

\citet{maini2019adversarial} report results for only MSD-50 in their paper. We produce MSD-$K$ results using their publicly available code. While reducing the number of steps in MSD, we \emph{appropriately} increase the step size $\alpha$ for the attack. For MSD-50, \citet{maini2019adversarial} used $\alpha=(0.003,0.05,1.0)$ for $(\ell_\infty, \ell_2, \ell_1)$ perturbations, respectively. We proportionately increase the step size to $\alpha=(0.005,0.084,1.68)$ and $\alpha=(0.0075,0.125,2.5)$ for MSD-30 and MSD-20, respectively. 

For MSD-10 and MSD-5, we choose $\alpha=(0.0075,0.125,2.5)$, since we found that further increasing the step size $\alpha$  lead to \emph{lower} final adversarial accuracy. 

Other than the step-size, we do \emph{not} make any change to the original code by \citet{maini2019adversarial}. 

{\textbf{AVG-$K$~\cite{tramer2019adversarial} experiments for $K\in\{ 30, 20, 10, 5 \}$:}}

For AVG-50, we use the publicly available model provided by \citet{maini2019adversarial}. We produce AVG-$K$ results using the \citet{maini2019adversarial} code. When reducing the number of steps, we appropriately increase the step size $\alpha$ for $\ell_{\infty}$ and $\ell_2$ perturbations. Increasing the step size for $\ell_1$ perturbations resulted in significantly lower $\mathcal{A}^{(U)}_\text{adv}$, and thus $\alpha$ for $\ell_1$ perturbations was kept constant while reducing the number of steps. For AVG-50, \citet{maini2019adversarial} used $\alpha=(0.003,0.05,1.0)$ for $(\ell_\infty, \ell_2, \ell_1)$ perturbations, respectively. We increase the $\ell_{\infty}$ and $\ell_2$ step sizes to set $\alpha=(0.005,0.084,1.0)$ and $\alpha=(0.0075,0.125,1.0)$ for AVG-30 and AVG-20 respectively.

As with MSD, we do not further increase the step size $\alpha$ for AVG-10, AVG-5, and instead choose $\alpha=(0.0075,0.125,1.0)$. Even here, we found that increasing the step size for $\ell_1$ perturbations results in lower $\mathcal{A}^{(U)}_\text{adv}$. For AVG-2, we increase the step size for all perturbations to $\alpha=(0.024,0.4,8)$.

{\textbf{PAT~\citep{laidlaw2021perceptual} on CIFAR-10:}}

For comparisons with \citet{laidlaw2021perceptual}, we evaluate their publicly available {self-bounded} ResNet-50 model. %

\subsection{Details about SNAP}

\subsubsection{Distribution Update Epoch}

In the SNAP distribution update epoch (see Algorithm~1 in the main text), we employ $\ell_2$ norm-bounded PGD attack to compute perturbation vectors $\bm{\eta}$. We use only  20\% of the training data, which is randomly selected during every SNAP update epoch. Recall that normalized root mean squared projections of $\bm{\eta}$ dictate the updated noise variances (Eq.~(3) in the main text). In the following we provide more details specific to CIFAR-10 and ImageNet data:

CIFAR-10: we employ 10 step $\ell_2$-PGD attack with $\epsilon=1.8$ and $N_0=4$. 

ImageNet: we employ 4 step $\ell_2$-PGD attack with $\epsilon=4.0$ and $N_0=1$.

Note that $\ell_2$ norm bound $\epsilon$ for the PGD attack here does not play any role, since $\bm{\eta}$ perturbation projections are normalized.

\subsubsection{Noise variance initialization in SNAP}

In SNAP, we initialize the noise variances to be uniform across all dimensions. Specifically, in Algorithm~1 in the main text, $\Sigma_0=\text{Diag}\bigg[\sqrt{\frac{P_\text{noise}}{D}},\dots,\sqrt{\frac{P_\text{noise}}{D}}\bigg]$ for a given value of $P_\text{noise}$. 

 \section{Accompanying Code and Pretrained Models}

As a part of this appendix, we share our code to reproduce PGD+SNAP and TRADES+SNAP results on CIFAR-10 (Table~2 in the main text) and FreeAdv+SNAP results on ImageNet (Table~4 in the main text). We also share corresponding pretrained models to facilitate quick reproduction of our results. Code and models are available at link: \href{https://github.com/adpatil2/SNAP}{https://github.com/adpatil2/SNAP}

\end{appendices}

%% file: main.bbl
\medskip


{
\small
\begin{thebibliography}{48}
\providecommand{\natexlab}[1]{#1}
\providecommand{\url}[1]{\texttt{#1}}
\expandafter\ifx\csname urlstyle\endcsname\relax
  \providecommand{\doi}[1]{doi: #1}\else
  \providecommand{\doi}{doi: \begingroup \urlstyle{rm}\Url}\fi

\bibitem[Andriushchenko et~al.(2020)Andriushchenko, Croce, Flammarion, and
  Hein]{andriushchenko2020square}
Andriushchenko, M., Croce, F., Flammarion, N., and Hein, M.
\newblock Square attack: a query-efficient black-box adversarial attack via
  random search.
\newblock In \emph{European Conference on Computer Vision}, pp.\  484--501.
  Springer, 2020.

\bibitem[Bhattad et~al.(2019)Bhattad, Chong, Liang, Li, and
  Forsyth]{bhattad2019unrestricted}
Bhattad, A., Chong, M.~J., Liang, K., Li, B., and Forsyth, D.~A.
\newblock Unrestricted adversarial examples via semantic manipulation.
\newblock \emph{arXiv preprint arXiv:1904.06347}, 2019.

\bibitem[Brendel et~al.(2018)Brendel, Rauber, and Bethge]{brendel2018decision}
Brendel, W., Rauber, J., and Bethge, M.
\newblock Decision-based adversarial attacks: Reliable attacks against
  black-box machine learning models.
\newblock In \emph{International Conference on Learning Representations}, 2018.

\bibitem[Chen et~al.(2018)Chen, Sharma, Zhang, Yi, and Hsieh]{chen2018ead}
Chen, P.-Y., Sharma, Y., Zhang, H., Yi, J., and Hsieh, C.-J.
\newblock Ead: elastic-net attacks to deep neural networks via adversarial
  examples.
\newblock In \emph{Thirty-second AAAI conference on artificial intelligence},
  2018.

\bibitem[Cohen et~al.(2019)Cohen, Rosenfeld, and Kolter]{cohen2019certified}
Cohen, J., Rosenfeld, E., and Kolter, Z.
\newblock Certified adversarial robustness via randomized smoothing.
\newblock In \emph{International Conference on Machine Learning (ICML)}, 2019.

\bibitem[Dezfooli et~al.(2018)Dezfooli, Fawzi, Fawzi, Frossard, and
  Soatto]{dezfooli2018robustness}
Dezfooli, S. M.~M., Fawzi, A., Fawzi, O., Frossard, P., and Soatto, S.
\newblock Robustness of classifiers to universal pertur-bations: A geometric
  perspective.
\newblock In \emph{International Conference on Learning Representations
  (ICLR)}, 2018.

\bibitem[Engstrom et~al.(2019)Engstrom, Tran, Tsipras, Schmidt, and
  Madry]{engstrom2019exploring}
Engstrom, L., Tran, B., Tsipras, D., Schmidt, L., and Madry, A.
\newblock Exploring the landscape of spatial robustness.
\newblock In \emph{International Conference on Machine Learning}, pp.\
  1802--1811. PMLR, 2019.

\bibitem[Gilmer et~al.(2019)Gilmer, Ford, Carlini, and
  Cubuk]{gilmer2019adversarial}
Gilmer, J., Ford, N., Carlini, N., and Cubuk, E.
\newblock Adversarial examples are a natural consequence of test error in
  noise.
\newblock In \emph{International Conference on Machine Learning}, pp.\
  2280--2289, 2019.

\bibitem[Gowal et~al.(2020)Gowal, Qin, Uesato, Mann, and
  Kohli]{gowal2020uncovering}
Gowal, S., Qin, C., Uesato, J., Mann, T., and Kohli, P.
\newblock Uncovering the limits of adversarial training against norm-bounded
  adversarial examples.
\newblock \emph{arXiv preprint arXiv:2010.03593}, 2020.

\bibitem[Gui et~al.(2019)Gui, Wang, Yu, Yang, Wang, and Liu]{gui2019model}
Gui, S., Wang, H., Yu, C., Yang, H., Wang, Z., and Liu, J.
\newblock Model compression with adversarial robustness: A unified optimization
  framework.
\newblock \emph{arXiv preprint arXiv:1902.03538}, 2019.

\bibitem[Guo et~al.(2020)Guo, Yang, Xu, Liu, and Lin]{guo2020meets}
Guo, M., Yang, Y., Xu, R., Liu, Z., and Lin, D.
\newblock When nas meets robustness: In search of robust architectures against
  adversarial attacks.
\newblock In \emph{Proceedings of the IEEE/CVF Conference on Computer Vision
  and Pattern Recognition}, pp.\  631--640, 2020.

\bibitem[He et~al.(2019)He, Rakin, and Fan]{he2019parametric}
He, Z., Rakin, A.~S., and Fan, D.
\newblock Parametric noise injection: Trainable randomness to improve deep
  neural network robustness against adversarial attack.
\newblock In \emph{Proceedings of the IEEE Conference on Computer Vision and
  Pattern Recognition (CVPR)}, 2019.

\bibitem[Hendrycks \& Dietterich(2018)Hendrycks and
  Dietterich]{hendrycks2018benchmarking}
Hendrycks, D. and Dietterich, T.
\newblock Benchmarking neural network robustness to common corruptions and
  perturbations.
\newblock In \emph{International Conference on Learning Representations}, 2018.

\bibitem[Hu et~al.(2020)Hu, Chen, Wang, and Wang]{hu2020triple}
Hu, T.-K., Chen, T., Wang, H., and Wang, Z.
\newblock Triple wins: Boosting accuracy, robustness and efficiency together by
  enabling input-adaptive inference.
\newblock \emph{arXiv preprint arXiv:2002.10025}, 2020.

\bibitem[Ilyas et~al.(2019)Ilyas, Santurkar, Tsipras, Engstrom, Tran, and
  Madry]{ilyas2019adversarial}
Ilyas, A., Santurkar, S., Tsipras, D., Engstrom, L., Tran, B., and Madry, A.
\newblock Adversarial examples are not bugs, they are features.
\newblock \emph{arXiv preprint arXiv:1905.02175}, 2019.

\bibitem[Jordan et~al.(2019)Jordan, Manoj, Goel, and
  Dimakis]{jordan2019quantifying}
Jordan, M., Manoj, N., Goel, S., and Dimakis, A.~G.
\newblock Quantifying perceptual distortion of adversarial examples.
\newblock \emph{arXiv preprint arXiv:1902.08265}, 2019.

\bibitem[Kang et~al.(2019)Kang, Sun, Brown, Hendrycks, and
  Steinhardt]{kang2019transfer}
Kang, D., Sun, Y., Brown, T., Hendrycks, D., and Steinhardt, J.
\newblock Transfer of adversarial robustness between perturbation types.
\newblock \emph{arXiv preprint arXiv:1905.01034}, 2019.

\bibitem[Laidlaw \& Feizi(2019)Laidlaw and Feizi]{laidlaw2019functional}
Laidlaw, C. and Feizi, S.
\newblock Functional adversarial attacks.
\newblock \emph{Advances in Neural Information Processing Systems}, 2019.

\bibitem[Laidlaw et~al.(2018)Laidlaw, Singla, and Feizi]{laidlaw2021perceptual}
Laidlaw, C., Singla, S., and Feizi, S.
\newblock Perceptual adversarial robustness: Defense against unseen threat
  models.
\newblock \emph{International Conference on Learning Representations (ICLR)},
  2018.

\bibitem[Li et~al.(2019)Li, Chen, Wang, and Duke]{li2019certified}
Li, B., Chen, C., Wang, W., and Duke, L.~C.
\newblock Certified adversarial robustness with addition gaussian noise.
\newblock \emph{Neural Information Processing Systems (NeurIPS)}, 2019.

\bibitem[Madaan et~al.(2020)Madaan, Shin, and Hwang]{madaan2020learning}
Madaan, D., Shin, J., and Hwang, S.~J.
\newblock Learning to generate noise for robustness against multiple
  perturbations.
\newblock \emph{arXiv preprint arXiv:2006.12135}, 2020.

\bibitem[Madry et~al.(2018)Madry, Makelov, Schmidt, Tsipras, and
  Vladu]{madry2018towards}
Madry, A., Makelov, A., Schmidt, L., Tsipras, D., and Vladu, A.
\newblock Towards deep learning models resistant to adversarial attacks.
\newblock \emph{International Conference on Learning Representations (ICLR)},
  2018.

\bibitem[Maini et~al.(2020)Maini, Wong, and Kolter]{maini2019adversarial}
Maini, P., Wong, E., and Kolter, J.~Z.
\newblock Adversarial robustness against the union of multiple perturbation
  models.
\newblock In \emph{International Conference on Machine Learning (ICML)}, 2020.

\bibitem[Moosavi-Dezfooli et~al.(2016)Moosavi-Dezfooli, Fawzi, and
  Frossard]{moosavi2016deepfool}
Moosavi-Dezfooli, S.-M., Fawzi, A., and Frossard, P.
\newblock Deepfool: a simple and accurate method to fool deep neural networks.
\newblock In \emph{Proceedings of the IEEE conference on computer vision and
  pattern recognition (CVPR)}, 2016.

\bibitem[Moosavi-Dezfooli et~al.(2019)Moosavi-Dezfooli, Fawzi, Uesato, and
  Frossard]{moosavi2019robustness}
Moosavi-Dezfooli, S.-M., Fawzi, A., Uesato, J., and Frossard, P.
\newblock Robustness via curvature regularization, and vice versa.
\newblock In \emph{Proceedings of the IEEE Conference on Computer Vision and
  Pattern Recognition (CVPR)}, 2019.

\bibitem[Pinot et~al.(2019)Pinot, Meunier, Araujo, Kashima, Yger, Gouy-Pailler,
  and Atif]{pinot2019theoretical}
Pinot, R., Meunier, L., Araujo, A., Kashima, H., Yger, F., Gouy-Pailler, C.,
  and Atif, J.
\newblock Theoretical evidence for adversarial robustness through
  randomization: the case of the exponential family.
\newblock In \emph{Advances in Neural Information Processing Systems}, 2019.

\bibitem[Pinot et~al.(2020)Pinot, Ettedgui, Rizk, Chevaleyre, and
  Atif]{pinot2020randomization}
Pinot, R., Ettedgui, R., Rizk, G., Chevaleyre, Y., and Atif, J.
\newblock Randomization matters. how to defend against strong adversarial
  attacks.
\newblock In \emph{International Conference on Machine Learning (ICML)}, 2020.

\bibitem[Rauber et~al.(2020)Rauber, Zimmermann, Bethge, and
  Brendel]{rauber2017foolboxnative}
Rauber, J., Zimmermann, R., Bethge, M., and Brendel, W.
\newblock Foolbox native: Fast adversarial attacks to benchmark the robustness
  of machine learning models in pytorch, tensorflow, and jax.
\newblock \emph{Journal of Open Source Software}, 5\penalty0 (53):\penalty0
  2607, 2020.
\newblock \doi{10.21105/joss.02607}.
\newblock URL \url{https://doi.org/10.21105/joss.02607}.

\bibitem[Rebuffi et~al.(2021)Rebuffi, Gowal, Calian, Stimberg, Wiles, and
  Mann]{rebuffi2021fixing}
Rebuffi, S.-A., Gowal, S., Calian, D.~A., Stimberg, F., Wiles, O., and Mann, T.
\newblock Fixing data augmentation to improve adversarial robustness.
\newblock \emph{arXiv preprint arXiv:2103.01946}, 2021.

\bibitem[Rice et~al.(2020)Rice, Wong, and Kolter]{rice2020overfitting}
Rice, L., Wong, E., and Kolter, Z.
\newblock Overfitting in adversarially robust deep learning.
\newblock In \emph{International Conference on Machine Learning}, pp.\
  8093--8104. PMLR, 2020.

\bibitem[Rony et~al.(2019)Rony, Hafemann, Oliveira, Ayed, Sabourin, and
  Granger]{rony2019decoupling}
Rony, J., Hafemann, L.~G., Oliveira, L.~S., Ayed, I.~B., Sabourin, R., and
  Granger, E.
\newblock Decoupling direction and norm for efficient gradient-based l2
  adversarial attacks and defenses.
\newblock In \emph{Proceedings of the IEEE/CVF Conference on Computer Vision
  and Pattern Recognition}, pp.\  4322--4330, 2019.

\bibitem[Rusak et~al.(2020)Rusak, Schott, Zimmermann, Bitterwolf, Bringmann,
  Bethge, and Brendel]{rusak2020simple}
Rusak, E., Schott, L., Zimmermann, R.~S., Bitterwolf, J., Bringmann, O.,
  Bethge, M., and Brendel, W.
\newblock A simple way to make neural networks robust against diverse image
  corruptions.
\newblock In \emph{European Conference on Computer Vision}, pp.\  53--69.
  Springer, 2020.

\bibitem[Salman et~al.(2019)Salman, Li, Razenshteyn, Zhang, Zhang, Bubeck, and
  Yang]{salman2019provably}
Salman, H., Li, J., Razenshteyn, I., Zhang, P., Zhang, H., Bubeck, S., and
  Yang, G.
\newblock Provably robust deep learning via adversarially trained smoothed
  classifiers.
\newblock In \emph{Advances in Neural Information Processing Systems}, pp.\
  11289--11300, 2019.

\bibitem[Santurkar et~al.(2019)Santurkar, Ilyas, Tsipras, Engstrom, Tran, and
  Madry]{santurkar2019image}
Santurkar, S., Ilyas, A., Tsipras, D., Engstrom, L., Tran, B., and Madry, A.
\newblock Image synthesis with a single (robust) classifier.
\newblock In \emph{Advances in Neural Information Processing Systems}, pp.\
  1262--1273, 2019.

\bibitem[Shafahi et~al.(2019)Shafahi, Najibi, Ghiasi, Xu, Dickerson, Studer,
  Davis, Taylor, and Goldstein]{shafahi2019adversarial}
Shafahi, A., Najibi, M., Ghiasi, A., Xu, Z., Dickerson, J., Studer, C., Davis,
  L.~S., Taylor, G., and Goldstein, T.
\newblock Adversarial training for free!
\newblock \emph{Advances in Neural Information Processing Systems (NeurIPS)},
  2019.

\bibitem[Stutz et~al.(2020)Stutz, Hein, and Schiele]{stutz2020confidence}
Stutz, D., Hein, M., and Schiele, B.
\newblock Confidence-calibrated adversarial training: Generalizing to unseen
  attacks.
\newblock In \emph{International Conference on Machine Learning}, pp.\
  9155--9166. PMLR, 2020.

\bibitem[Tram{\`e}r \& Boneh(2019)Tram{\`e}r and Boneh]{tramer2019adversarial}
Tram{\`e}r, F. and Boneh, D.
\newblock Adversarial training and robustness for multiple perturbations.
\newblock In \emph{Advances in Neural Information Processing Systems}, pp.\
  5858--5868, 2019.

\bibitem[Tramer et~al.(2020)Tramer, Carlini, Brendel, and
  Madry]{tramer2020adaptive}
Tramer, F., Carlini, N., Brendel, W., and Madry, A.
\newblock On adaptive attacks to adversarial example defenses.
\newblock \emph{arXiv preprint arXiv:2002.08347}, 2020.

\bibitem[Vivek \& Babu(2020)Vivek and Babu]{vivek2020single}
Vivek, B. and Babu, R.~V.
\newblock Single-step adversarial training with dropout scheduling.
\newblock In \emph{2020 IEEE/CVF Conference on Computer Vision and Pattern
  Recognition (CVPR)}, pp.\  947--956. IEEE, 2020.

\bibitem[Wong et~al.(2020)Wong, Rice, and Kolter]{wong2020fast}
Wong, E., Rice, L., and Kolter, J.~Z.
\newblock Fast is better than free: Revisiting adversarial training.
\newblock In \emph{International Conference on Machine Learning (ICLR)}, 2020.

\bibitem[Xiao et~al.(2018)Xiao, Zhu, Li, He, Liu, and Song]{xiao2018spatially}
Xiao, C., Zhu, J.-Y., Li, B., He, W., Liu, M., and Song, D.
\newblock Spatially transformed adversarial examples.
\newblock In \emph{International Conference on Learning Representations}, 2018.

\bibitem[Xie \& Yuille(2020)Xie and Yuille]{xie2020intriguing}
Xie, C. and Yuille, A.
\newblock Intriguing properties of adversarial training at scale.
\newblock In \emph{International Conference on Learning Representations}, 2020.

\bibitem[Yang et~al.(2020{\natexlab{a}})Yang, Duan, Hu, Salman, Razenshteyn,
  and Li]{yang2020randomized}
Yang, G., Duan, T., Hu, E., Salman, H., Razenshteyn, I., and Li, J.
\newblock Randomized smoothing of all shapes and sizes.
\newblock \emph{International Conference on Machine Learning (ICML)},
  2020{\natexlab{a}}.

\bibitem[Yang et~al.(2020{\natexlab{b}})Yang, Rashtchian, Zhang, Salakhutdinov,
  and Chaudhuri]{yang2020closer}
Yang, Y.-Y., Rashtchian, C., Zhang, H., Salakhutdinov, R., and Chaudhuri, K.
\newblock A closer look at accuracy vs. robustness.
\newblock \emph{Advances in Neural Information Processing Systems}, 33,
  2020{\natexlab{b}}.

\bibitem[Zhang et~al.(2019{\natexlab{a}})Zhang, Zhang, Lu, Zhu, and
  Dong]{zhang2019you}
Zhang, D., Zhang, T., Lu, Y., Zhu, Z., and Dong, B.
\newblock You only propagate once: Accelerating adversarial training via
  maximal principle.
\newblock \emph{arXiv preprint arXiv:1905.00877}, 2019{\natexlab{a}}.

\bibitem[Zhang et~al.(2019{\natexlab{b}})Zhang, Yu, Jiao, Xing, El~Ghaoui, and
  Jordan]{zhang2019theoretically}
Zhang, H., Yu, Y., Jiao, J., Xing, E., El~Ghaoui, L., and Jordan, M.
\newblock Theoretically principled trade-off between robustness and accuracy.
\newblock In \emph{International Conference on Machine Learning (ICML)},
  2019{\natexlab{b}}.

\bibitem[Zhang et~al.(2020)Zhang, Xu, Han, Niu, Cui, Sugiyama, and
  Kankanhalli]{zhang2020attacks}
Zhang, J., Xu, X., Han, B., Niu, G., Cui, L., Sugiyama, M., and Kankanhalli, M.
\newblock Attacks which do not kill training make adversarial learning
  stronger.
\newblock In \emph{International Conference on Machine Learning}, pp.\
  11278--11287. PMLR, 2020.

\bibitem[Zheng et~al.(2020)Zheng, Zhang, Gu, Lee, and
  Prakash]{zheng2020efficient}
Zheng, H., Zhang, Z., Gu, J., Lee, H., and Prakash, A.
\newblock Efficient adversarial training with transferable adversarial
  examples.
\newblock In \emph{Proceedings of the IEEE/CVF Conference on Computer Vision
  and Pattern Recognition}, pp.\  1181--1190, 2020.

\end{thebibliography}

}
